\documentclass[lettersize,journal]{IEEEtran}
\usepackage{amsmath,amsfonts}
\usepackage{algorithmic}
\usepackage{array}
\usepackage{textcomp}
\usepackage{stfloats}
\usepackage{url}
\usepackage{verbatim}
\usepackage{graphicx}
\usepackage{subfigure}
\usepackage[lined,ruled,noend]{algorithm2e}
\usepackage{tabularx}
\usepackage{multicol}
\usepackage{multirow}
\usepackage{booktabs} %
\usepackage{graphicx}
\usepackage{xcolor}
\usepackage{diagbox}
\usepackage{threeparttable}
\usepackage{amsmath}
\usepackage{multirow}
\usepackage[normalem]{ulem}
\usepackage[version=4]{mhchem}
\usepackage{svg}
\usepackage{ulem}
\usepackage{makecell}

\usepackage{mathrsfs}
\usepackage[mathscr]{euscript}
\usepackage{circledsteps}
\usepackage{ragged2e}
\usepackage{float}
\usepackage{tabularx}
\def\BibTeX{{\rm B\kern-.05em{\sc i\kern-.025em b}\kern-.08em
    T\kern-.1667em\lower.7ex\hbox{E}\kern-.125emX}}
\usepackage{balance}
\usepackage{xcolor,colortbl}

\usepackage{amsmath}

\begin{document}
\title{Mind the Gap: Learning Modality-Agnostic Representations with a Cross-Modality UNet}

\author{Xin~Niu,
        Enyi~Li,
        Jinchao~Liu,
        Yan~Wang,
        Margarita~Osadchy
        and~Yongchun~Fang

\thanks{Xin Niu, Enyi Li, Jinchao Liu and Yongchun Fang are with Tianjin Key Laboratory of Intelligent Robotics, College of Artificial Intelligence, Nankai University, China, and also with Engineering Research Center of Trusted Behavior Intelligence, Ministry of Education, Nankai University, China. e-mail: \{niuxin,lienyi\}@mail.nankai.edu.cn, \{liujinchao,fangyc\}@nankai.edu.cn (\textit{Xin Niu and Enyi Li are co-first authors}) (\textit{Corresponding author: Jinchao Liu})}
\thanks{Yan Wang is with VisionMetric Ltd, Canterbury, Kent, UK. e-mail: yanwangnott@gmail.com}
\thanks{Margarita Osadchy is with the Department of Computer Science, Haifa University, Israel. e-mail: rita@cs.haifa.ac.il}

\thanks{\footnotesize{\copyright 2024 IEEE. Personal use of this material is permitted.  Permission from IEEE must be obtained for all other uses, in any current or future media, including reprinting/republishing this material for advertising or promotional purposes, creating new collective works, for resale or redistribution to servers or lists, or reuse of any copyrighted component of this work in other works.}}

\thanks{This work has been published in the IEEE Transactions on Image Processing, vol. 33, pp. 655-670, 2024, doi: 10.1109/TIP.2023.3348656.}
}

\markboth{IEEE Transactions on Image Processing}%
{How to Use the IEEEtran \LaTeX \ Templates}

\maketitle

\begin{abstract}

Cross-modality recognition has many important applications in science, law enforcement and entertainment. 
Popular methods to bridge the modality gap include reducing the distributional differences of representations of different modalities, learning indistinguishable representations or explicit modality transfer. The first two approaches suffer from the loss of discriminant information while removing the modality-specific variations. The third one heavily relies on the successful modality transfer, could face catastrophic performance drop when explicit modality transfers are not possible or difficult. To tackle this problem, we proposed a compact encoder-decoder neural module (cmUNet) to learn modality-agnostic representations while retaining identity-related information. This is achieved through cross-modality transformation and in-modality reconstruction, enhanced by an adversarial/perceptual loss which encourages indistinguishability of representations in the original sample space. 
For cross-modality matching, we propose MarrNet where cmUNet is connected to a standard feature extraction network which takes as inputs the modality-agnostic representations and outputs similarity scores for matching. 
We validated our method on five challenging tasks, namely Raman-infrared spectrum matching, cross-modality person re-identification and heterogeneous (photo-sketch, visible-near infrared and visible-thermal) face recognition, where MarrNet showed superior performance compared to state-of-the-art methods. 
Furthermore, it is observed that a cross-modality matching method could be biased to extract discriminant information from partial or even wrong regions, due to incompetence of dealing with modality gaps, which subsequently leads to poor generalization. We show that robustness to occlusions can be an indicator of whether a method can well bridge the modality gap. This, to our knowledge, has been largely neglected in the previous works. Our experiments demonstrated that MarrNet exhibited excellent robustness against disguises and occlusions, and outperformed existing methods with a large margin ($>10\%$). The proposed cmUNet is a meta-approach and can be used as a building block for various applications.

\end{abstract}


\section{Introduction}

Cross-modality recognition has many important applications and it also serves as an excellent platform for exploring advanced representation learning frameworks. This is because matching inputs from different modalities requires learning purely semantic features that do not include modality-specific information, which could interfere in matching.

\begin{figure}
  \centering
  \subfigure[Examples from the CUFSF Photo-Sketch Face Database]{
  \includegraphics[width=0.98\linewidth]{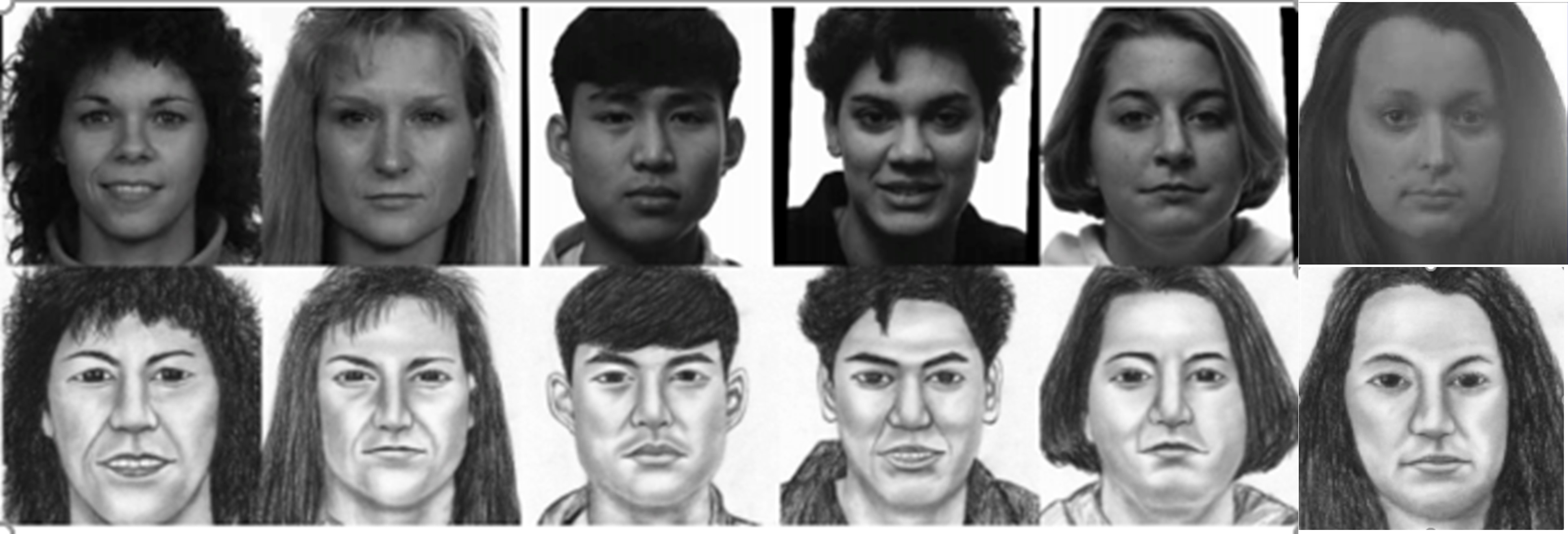}
  }
  \subfigure[``Thin-Ice Hypothesis": Bridging the modality gap matters]{
  \includegraphics[width=0.98\linewidth]{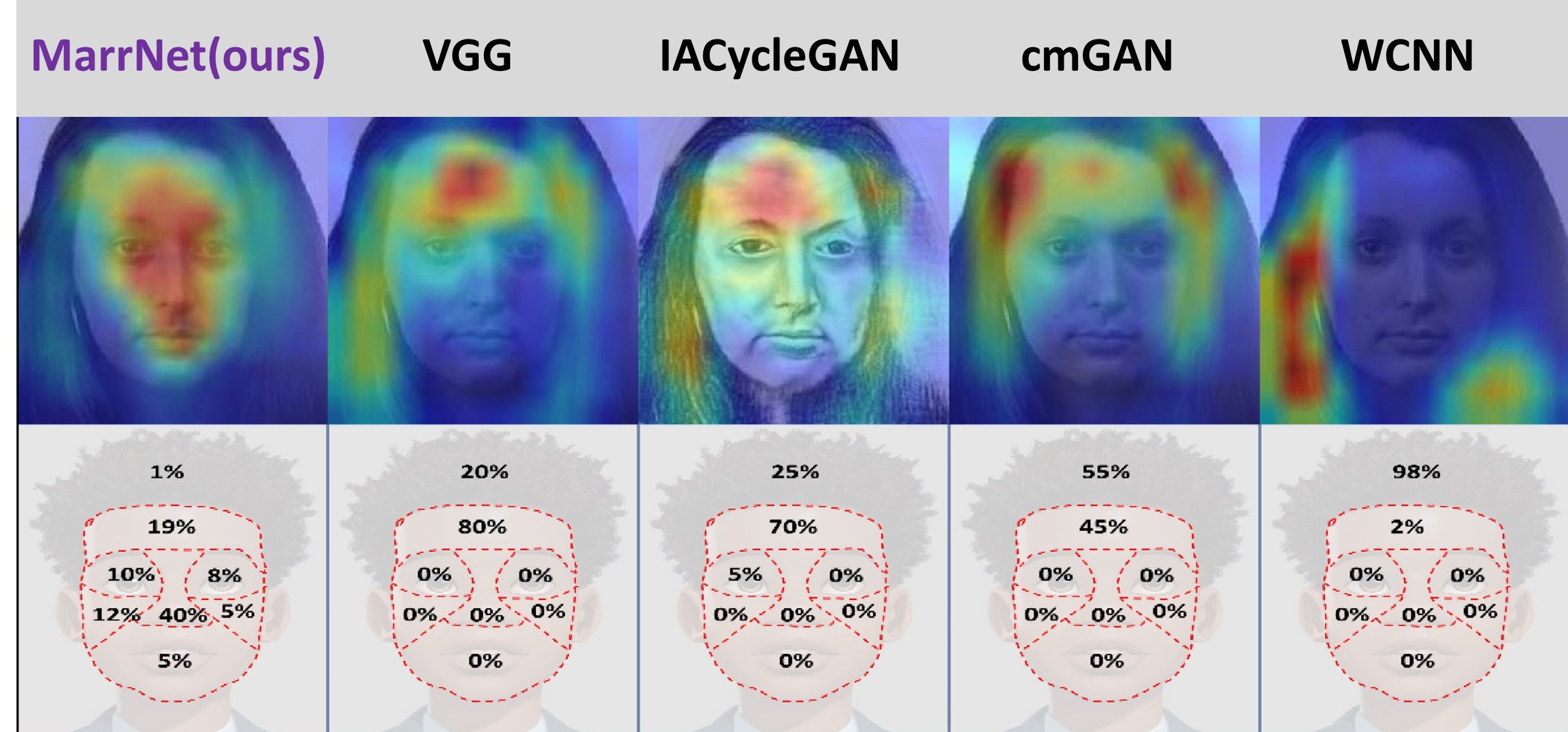}
  }
  \caption{Cross-modality Face Recognition. (a) Examples of faces (top row) and sketches (bottom row) from the dataset CUFSF. (b) Saliency maps of the compared methods, and relative importance of different regions of face images contributing to decision making. It is observed that representations learned by existing methods seemed to wrongly-focused on areas, in which the modality gaps were small, due to incompetence of dealing with modality gaps ("Thin-Ice Hypothesis"). We demonstrated that the proposed MarrNet was able to bridge the modality gap well and focus on the true discriminant regions.}
    \label{fig_first_impression}
\end{figure}

There are abundant practical applications of cross-modality matching. Popular ones that grew into their own research area are heterogeneous face recognition involving modalities of photo vs sketch, visible vs near infrared and visible vs thermal~\cite{klare2010matching,7139092,8006255,1238414,1467376,4624272,7434636,fang2020identity,LIU2021107579,WCNN_TPAMI2019,DVGFace_TPAMI2022,IRIS_Thermal_Visible_Face_Database}, and cross-modality person re-identification~\cite{zhang2023diverse,zhang2021towards,zhang2022fmcnet}. Face sketches are used in entertainment and law enforcement~\cite{4624272}. In particular, in law enforcement, mug shots of suspects are not always available. Instead, face sketches according to the description from eyewitness or blurry surveillance videos. These sketches are then either circulated in public for identification or automatically matched to photos in the police’s criminal face databases. Examples of pairs of photo and sketches can be found in Fig.~\ref{fig_first_impression}(a). There are also applications in law enforcement that demand technologies of matching identities across face images acquired by different sensing instruments such as visible, near infrared or thermal cameras. Fig.~\ref{fig_datasets_examples_face} shows examples of all three tasks of cross-modality face recognition from the databases used in our study.

\begin{figure*}
  \centering
  \includegraphics[width=0.95\linewidth]{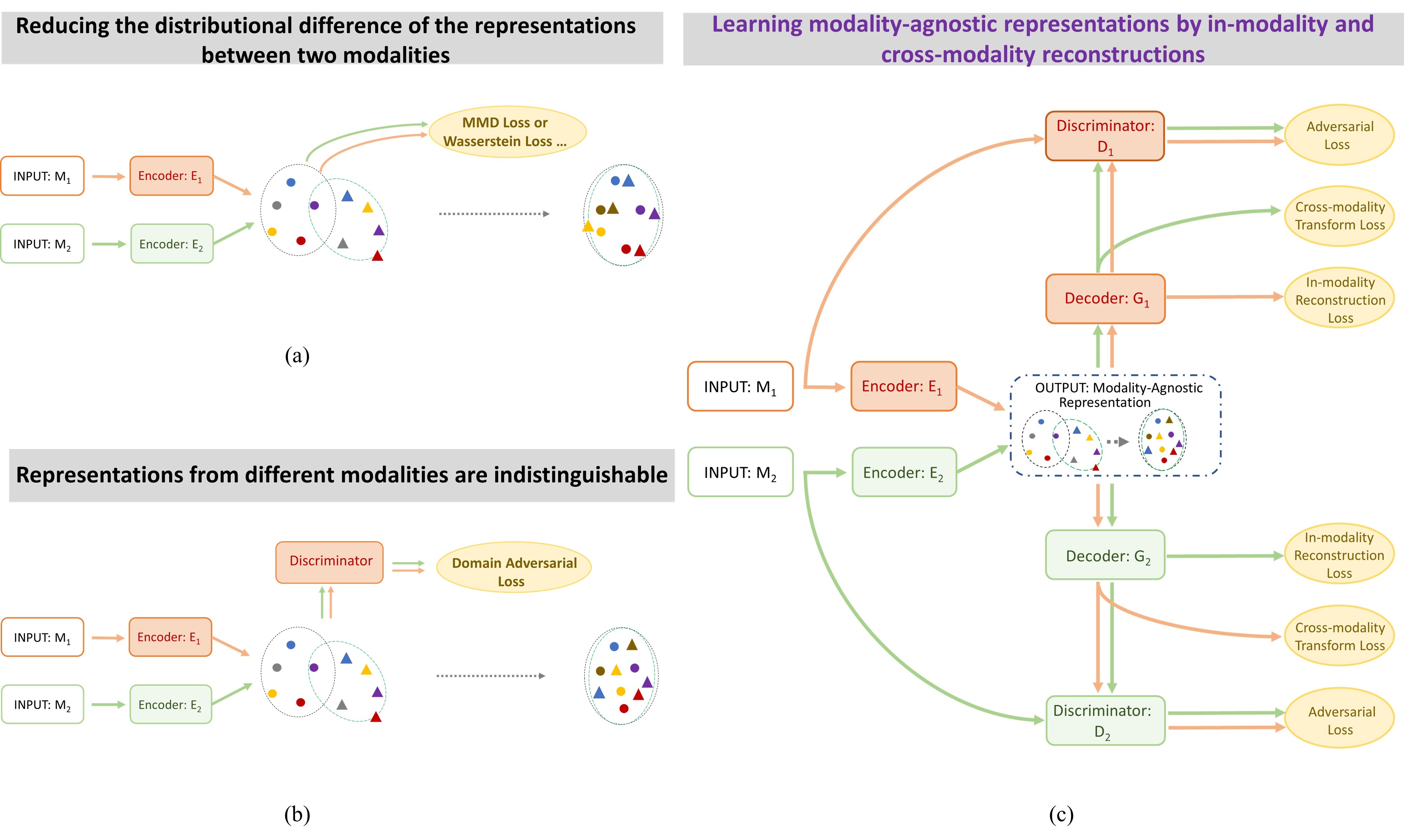}
  \caption{A graphical illustration of the proposed neural module \textit{cmUNet} which provides an alternative way of learning the modality-agnostic representation via self-supervised learning, compared to existing approaches most related to ours, including (a) reducing distributional difference of the representations between two modalities~\cite{WCNN_TPAMI2019}, with a cost of losing discriminate information~\cite{RethinkingMMD2023}. (b) forcing the learned representations from different modalities to be indistinguishable~\cite{tzeng2017adversarial}, where however potentially destroy all discriminant, beyond modality-related, information. (c) The proposed cmUNet performs both cross-modality transformation and in-modality reconstruction, where the former forces to remove the modality-related information while the latter encourages to retain the identity-related discriminant information, enhanced by an adversarial/perceptual loss.}
    \label{fig_cmUnet}
\end{figure*}

In addition to heterogeneous face recognition and person ReID, there are other applications of cross-modality matching. Here, we discuss vibrational spectroscopy -- matching between spectra of chemical substances obtained with different types of spectrometers, in particular Raman and Infrared.  Vibrational spectroscopy relies upon the inelastic scattering of monochromatic light, caused by interactions with molecular vibrations. A ``molecular fingerprint'' of a substance can therefore be obtained in the form of a spectrum comprising peaks that are characteristic of its chemical composition. It provides fast, non-contact, and non-destructive analysis and has a wide range of applications in a variety of industries, home land security, defense, medicine, ecology etc. Some examples are  mineral classification~\cite{liu2017,Liu2018}, chemical analysis~\cite{Fan2019}, pathogenic bacteria identification~\cite{Ho2019}, rapid detection of COVID-19 causative virus (SARS-CoV2)~\cite{RamanForCovid2021,RamanForCovid_Water_2022}, metabolite gradients monitoring~\cite{Lussier2019}, cytopathology~\cite{Kraub2018}, soil properties prediction~\cite{Liu2018,Padarian2019}, mine water inrush~\cite{Hu2019}. While deep networks introduced end-to-end approach to spectra matching that significantly improved over classical pipe-line methods~\cite{liu2017,LIU2019175,Ho2019}, the main challenge in using deep networks in this domain is the shortage of labeled data. Although, small datasets in different domains do exists. One way to resolve the shortage of training data is to provide tools that can operate over different types/modalities of spectra without any retraining.

Nowadays almost all cross-modality recognition systems are realized through deep neural networks, where representations are learned to bridge the gap between modalities or domains. Popular (meta-)methods include reducing distributional difference of the representations between two modalities using distances such as maximum mean discrepancy (MMD) and Wasserstein loss~\cite{WCNN_TPAMI2019, RethinkingMMD2023}, as shown in Fig.~\ref{fig_cmUnet}(a). However, these losses/constraints apply to the learned features directly and force the corresponding distributions to be close. As a result, not only the modality-related information is removed, but the discriminant information may also be damaged, as reported in~\cite{RethinkingMMD2023}.

Similar motivation implemented via adversarial training was suggested in~\cite{ganin2015unsupervised,motiian2017few,tzeng2017adversarial,volpi2018adversarial,ijcai2018p94}. In this line of work a discriminator was applied to the feature space to make the representations encoded from different modalities indistinguishable. While showing promising performance, this approach has an inherent drawback: forcing the representation in feature space to be \textit{indistinguishable} could potentially destroy all variability in learned representations, in particular, the identity information, causing degradation in the recognition accuracy. 

This observation motivated us to explore and design an alternative approach, which focuses on removing the modality-specific information \textit{only}, while retaining the identity-related discriminant information. Inspired by domain adversarial training and self-supervised learning, we present a cross-modality encoder-decoder neural module which performs both cross-modality transformation and in-modality reconstruction, guided by both reconstruction and adversarial/perceptual losses. The cross-modality transformation guided by an adversarial loss forces learned representations to be indistinguishable in the original sample space so as to remove modality-specific information. The in-modality reconstruction encourages the model to retain identity-related discriminant information to ensure successful reconstructions, as shown in Fig.~\ref{fig_cmUnet}(c). We named this neural module \textit{cmUNet} to emphasize there are shortcuts between encoders and decoders to facilitate the learning, see Fig.~\ref{fig_diagram_cmunet_spectrum} as an detailed example.

Another motivation for our approach is practical. We observed that common approaches for cross-modality face recognition benefit from generative models for e.g. mapping one modality to another with additional efforts of preserving person's identity~\cite{jing2019neural,jin2022deep}, or modelling the joint distribution of samples from both modalities (VIS and NIR face images) and producing a large number of augmented samples for training\cite{DVGFace_TPAMI2022}. Although intuitive, these methods suffer from several drawbacks. Firstly, domain transfer or joint modelling themselves is challenging and usually requires a large number of training data (e.g. the large face dataset \textit{MSCeleb-1M}). This limits the usage of these methods in small data applications. Secondly, image generation (one to one translation) is computationally intensive and basically unnecessary for recognition, since the generated images are subsequently used to learn semantic representations with no domain/modality information or bias. \textit{Is it possible to replace this complicated pipeline of domain transfer and representation learning with a single compact neural module?} This was also our motivation to design the cmUNet module and the MarrNet.

To this end, we propose a compact neural module for cross-modality matching which bridges the modality gap (remove modality-specific information only) and learns representations simultaneously so that the generation step in training\cite{fang2020identity,DVGFace_TPAMI2022} or inference\cite{fang2020identity} can be avoided. This is achieved by novel design shown in Fig.~\ref{fig_cmUnet}(c) which learns modality-agnostic representations by performing cross-modality transformation and in-modality reconstruction in training. For inference, all the auxiliary decoders and discriminators are discarded. 

\begin{figure*}
    \centering
    \includegraphics[width=\linewidth]{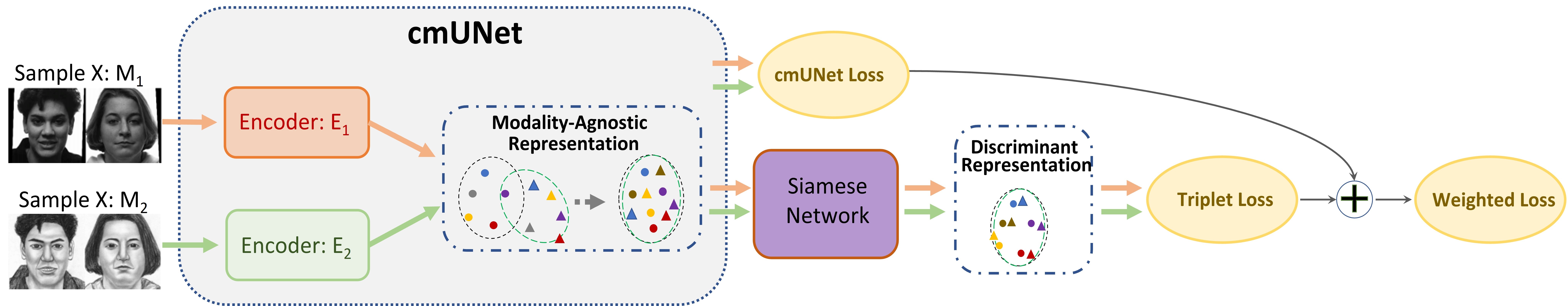}
    \caption{A graphical illustration of the proposed MarrNet for cross-modality matching. The network is designed such that, it first explicitly learns cross-modality representations prior to learning discriminant representations. This is realized by regularizing the modality-agnostic representations with a combination of cross-modality transformation and in-modality reconstruction. To keep this diagram uncluttered, only the encoders of the cmUNet are drawn. }
\label{fig_diagram_marrnet}
\end{figure*}

Our approach is a meta-approach and can be used as a building block for a variety of applications, especially small data tasks. Here we connect the cmUNet with a standard neural feature extractor for cross-modality matching (Fig.~\ref{fig_diagram_marrnet}). We demonstrate our approach on five challenging tasks, photo-sketch face recognition, visible-infrared face recognition, visible-thermal face recognition, visible-infrared person re-identification and
Raman-infrared spectrum matching. The only difference between the implementation of these tasks is the backbone architectures that are application tuned. Our generic approach applied to these distinct tasks achieve superior or competitive performance compared to the state-of-the-art methods, particularly our method on spectrum matching where transformation between modalities is of great difficulty, outperforms previous methods significantly. 

\vspace{0.3cm}
Our contributions are summarized as follows:
\begin{enumerate}
    \item We propose to learn modality-agnostic representations using a cross-modality encoder-decoder module (cmUNet) which performs cross-modality transformations and within-modality reconstruction to remove modality-specific information while retaining identity-related discriminant features. The cmUNet is a meta-approach and can be used for distinctly different tasks.
    
    \item For cross-modality matching, we propose MarrNet where cmUNet is connected to a standard feature extraction network which takes as inputs the modality-agnostic representations and outputs similarity scores for matching. We validated our method on five distinct challenging tasks, \textit{photo-sketch face recognition, visible-infrared face recognition, visible-thermal face recognition, visible-infrared person re-identification and Raman-infrared spectrum matching}, where the proposed method achieved superior performance against state-of-the-art methods.

    \item We propose ``thin-ice hypothesis'' to describe our finding that a cross-modality matching method could be biased to extract discriminant information from partial or even irrelevant regions, due to incompetence of dealing with modality gaps, which subsequently leads to poor generalization. We show that robustness to occlusions can be an indicator of whether a method can well bridge the modality gap. This, to our knowledge, has been largely neglected in the previous works. Our experiments demonstrate that MarrNet is highly robust to disguises and occlusions, and outperforms existing methods with a large margin ($>10\%$).
    
    \item We present a new problem \textit{Raman-infrared spectrum matching} and create a dataset \textit{cmRRUFF} which can be used as a benchmark for cross-modality representation learning~\footnote{We will release the dataset upon acceptance.}.    
    
\end{enumerate}

\section{Related Work}
There exist a number of applications that involve multiple modalities, such as heterogeneous face recognition\cite{fang2020identity}, image/video-text retrieval \cite{jiang2017deep, chen2021learning, bao2021dense, wang2022negative}, re-identification \cite{li2020infrared,fu2021cm,ijcai2018p94}, video captioning \cite{yang2021non,lin2021augmented}, audio-oriented machine comprehension \cite{huang2021audio}, multi-modality sentiment recognition. \cite{yu2021learning,zhang2021multi}. 

A gap between the modalities varies between applications in both size and the structure. For example, the gap between \textit{image/video} and \textit{text} is not only large, but also has little hierarchy. Alternatively, in some applications the gap could have a particular structure that could be exploited. Namely, in heterogeneous face recognition, the modality gap shows a clear hierarchy/structure, which includes a ``style change'' on top of the semantic features. Our approach is intended to utilize this structure by learning modality-agnostic representations prior to extracting semantic features, and targets heterogeneous face recognition, cross-modality person re-identification and cross-modality spectrum matching as typical tasks. 

To match samples across different modalities, early works for heterogeneous face recognition employed hand-crafted (cross-modality) features \cite{HandFeatureAIB2009,klare2010matching} involving features such as Local Binary Patterns (LBP), Gabor features and Difference of Gaussian (DoG). For instance, \cite{HandFeatureAIB2009}  combine DoG filtering and
multi-block LBP to encode NIR and VIS images. \cite{GoswamiHandFeatures2011} developed a recognition system based on Linear Discriminant Analysis on top of LBP features.

An alternative direction for both heterogeneous face recognition and person ReID was to map different modality inputs into a common latent space and apply matching in this space\cite{lin2006inter,FSSLHeRanPAMI2016,WCNN_TPAMI2019,zhang2023diverse,wei2021syncretic, fu2021cm,yang2022learning,chen2021neural}. More specifically, \cite{lin2006inter} proposed a common discriminant feature extraction (CDFE) approach. \cite{WCNN_TPAMI2019} designed separate layers to learn modality-invariant and -specific features and proposed to use Wasserstein distance to reduce the modality difference in the common shared latent space. \cite{chen2021neural} proposed modality-aware neural feature search to find a mapping to the latent space. 

Finally, there are methods that simply transformed images in one modality to another to form a homogeneous recognition problem including those achieving state-of-the-art performance in heterogeneous face recognition~\cite{1238414,1467376,4624272,zhang2015end,wang2018back,kazemi2018unsupervised,DVGFace_TPAMI2022} and in cross-modality person re-identification~\cite{zhang2021towards,wei2021syncretic}. For example, sketches were synthesized by Eigen transform~\cite{1238414}, local linear embedding~\cite{1467376} and Markov random fields~\cite{4624272}. Recent advances employed style transfer network architectures to map between modalities. \cite{zhang2015end} proposed an end-to-end photo-sketch generation model based on a 7-layer fully convolutional network (FCN) with a joint generative-discriminative optimization process. However, the transformed images were still not of the desired quality. \cite{wang2018back} used GAN to convert between sketch and photo. \cite{kazemi2018unsupervised} augmented CycleGan 
~\cite{wang2018high} with an additional geometry-discriminator to learn facial geometry. IACycleGAN~\cite{fang2020identity} added face recognition loss to the CycleGan in order to preserve the identity in the transformed image. The main problem in methods that employ an explicit transformation prior to recognition is that if the conversion produces poor results, the recognition accuracy degrades. \cite{DVGFace_TPAMI2022} modelled the joint distribution of samples from different modalities with a generative model which was then used to product augmented samples to train the cross-modality matching network. This method achieved remarkable performance, however required a large auxiliary dataset to train the generator. 

\section{MarrNet: Learning modality-agnostic representation with a cross-modality UNet}
It is beneficial to learn modality-agnostic representations for cross-modality recognition as modality-specific information could obscure the semantic information in the input. We characterise the  modality-agnostic representation as follows: for each pair of inputs from different modalities that correspond to the same sample, the representation must be homogeneous in the sense that all the information related to modality is removed.

As discussed previously, existing approaches such as reducing distributional distances or forcing representations to be indistinguishable in the feature space using an adversarial discriminator suffer from an inherent drawback: it removes modality-specific information often with a cost of losing discriminant information. 

We argue that representations are modality-agnostic if their decoding into sample space are indistinguishable regardless of the modality of the original samples. To learn such representations, we design a compact cross-modality encoder-decoder neural module which performs both cross-modality and within-modality reconstructions as shown in Fig.~\ref{fig_cmUnet}(c). Modality specific critics are added to check the indistinguishability of the learned representations decoded through decoders of different modalities. This neural module is named \textit{cmUNet} as there are also bridge connections between encoders and decoders to boost the learning. 

For cross-modality matching, we simply connect the cmUNet to a downstream feature extraction network to further learn discriminant representations, as shown in Fig.~\ref{fig_diagram_marrnet}. In other words, the modality-agnostic representation obtained by the encoder part of the cmUNet are passed to the Siamese network that operates on homogeneous inputs (in the feature space). When using a pretrained backbone, we can replicate the first few layers of the backbone as encoders, and create decoders correspondingly. The rest of the backbone is treated as the Siamese network (for discriminant feature learning). The proposed method is named as MarrNet which stands for Modality-Agnostic Representation Regularization for cross-modality matching.

\paragraph{Encoders} Two inputs $M_{1}$ and $M_{2}$ corresponding to two modalities of the same sample are first encoded by $E_{1}$ and $E_{2}$ respectively such that their outputs $E_{1}(M_{1})$ and $E_{2}(M_{2})$ are indistinguishable in terms of modality. In other words, we expect that they are transformed into a common feature space where they share the same ``latent'' modality. This will reduce the burden for the recognition network (The Siamese network $\textbf{S}$ in Fig.~\ref{fig_diagram_marrnet}) to bridge the gap between two modalities and to focus on learning discriminant information for final classification. 

\begin{figure}
  \centering
  \subfigure[``CUFSF Photo-Sketch Face Database"]{
  \includegraphics[width=0.98\linewidth]{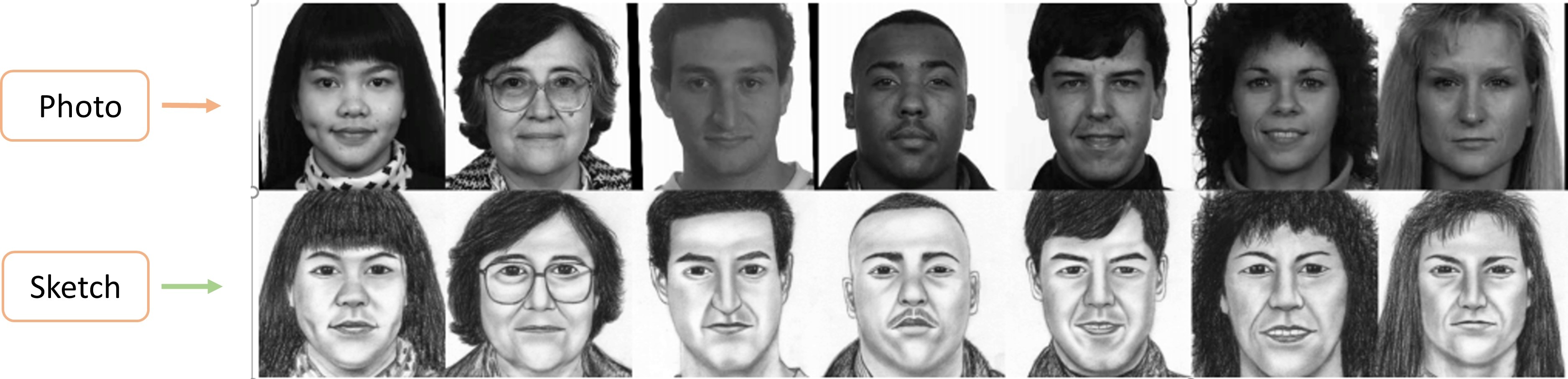}
  }
  \subfigure[``CASIA NIR-VIS 2.0 Face Database"]{
  \includegraphics[width=0.98\linewidth]{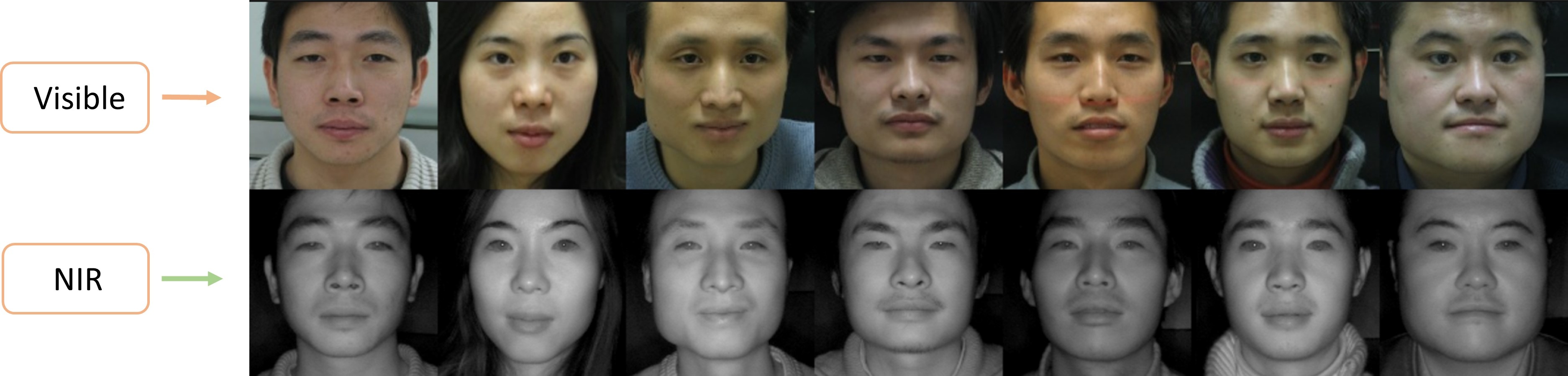}
  }
  \subfigure[``IRIS Thermal/Visible Face Database"]{
  \includegraphics[width=0.98\linewidth]{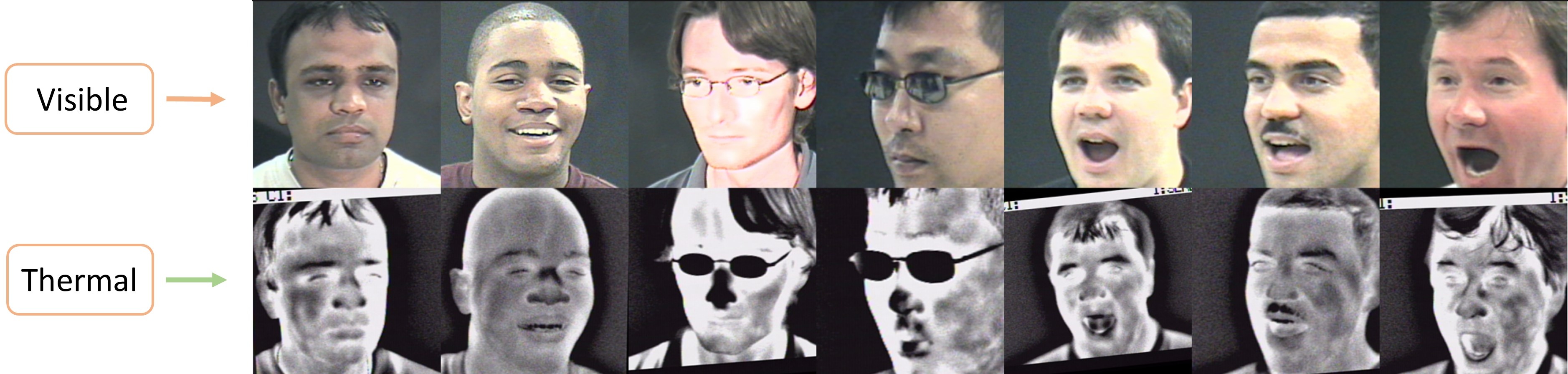}
  }
  \caption{Cross-modality face recognition. Examples of faces (top row) and sketches (bottom row) from the databases CUFSF, CASIA NIR-VIS 2.0 and IRIS Thermal/Visible Face Database. The modalities involved are photo vs sketch, visible vs near infrared and visible vs thermal respectively. The modality gap in CUFSF was introduced by cognitive drawing process of human artists, while the other two were caused by imaging sensors of different modalities.}
    \label{fig_datasets_examples_face}
\end{figure}

\paragraph{Decoders} We propose to decode the features $E_{1}(M_{1})$ and $E_{2}(M_{2})$ by two decoders $G_{i|i=1,2}$ and evaluate the reconstructed samples $G_{1}(E_{1}(M_{1}))$, $G_{2}(E_{2}(M_{2}))$ and the transformed samples $G_{1}(E_{2}(M_{2}))$, $G_{2}(E_{1}(M_{1}))$ qualitatively and quantitatively as descried below. 
\paragraph{Discriminators} We employ modality-specific discriminators $D_{i|i=1,2}$ in the setting of adversarial learning, each working against the corresponding decoder $G_i$ to evaluate domain's fidelity. 
\paragraph{Loss}
\noindent The outputs of the decoders $G_{i|i=1,2}$ that match the modality of the inputs are evaluated using the reconstruction loss:
\begin{align}
  L^{rec}_i=\left\| G_{i}(E_{i}(M_{i})) - M_{i} \right\|_{1}, \;\;i=\{1,2\}  
\end{align}

The outputs that are generated via the generator (decoder) from the opposite modality are evaluated using the cross-modality transform loss:
\begin{align}
L^{cross}_{ij}=\left\| G_{i}(E_{j}(M_{j})) - M_{i} \right\|_{1}, \;\; i,j=\{1,2\}; i\neq j
\end{align}

The transformed samples from the decoders are evaluated using the modality-aware discriminators compared to real samples, which contributes as an adversarial loss:
\begin{equation}
\begin{aligned}
 L^{adv}_{ij}&=\min_{G_i}\max_{D_i} \big[E_{M_i\sim P_{i}(M_i)}[\log(D_i(M_i)]\\& +
 E_{M_j\sim P_{j}(M_j)}[1-\log(D_i(G_i(E_j(M_j))))] \big] \\  
\end{aligned}
\end{equation}

where $i,j=\{1,2\};i\neq j$. $P_{i}$ and $P_{j}$ represent the distributions of samples of two modalities respectively. The adversarial loss regularizes the representations from different modalities, as a number of existing works showed that it is effective to remove domain specific information for domain adaptation applications~\cite{ganin2015unsupervised,motiian2017few,tzeng2017adversarial,volpi2018adversarial}. Note that we do not align the learned representations in the common feature space to force them to be indistinguishable, since doing so will not only remove domain specific features but also some of the class specific discriminant information, thus leading to a performance drop for the classification task.

\begin{figure}
  \centering
  \subfigure{
  \includegraphics[width=0.97\linewidth]{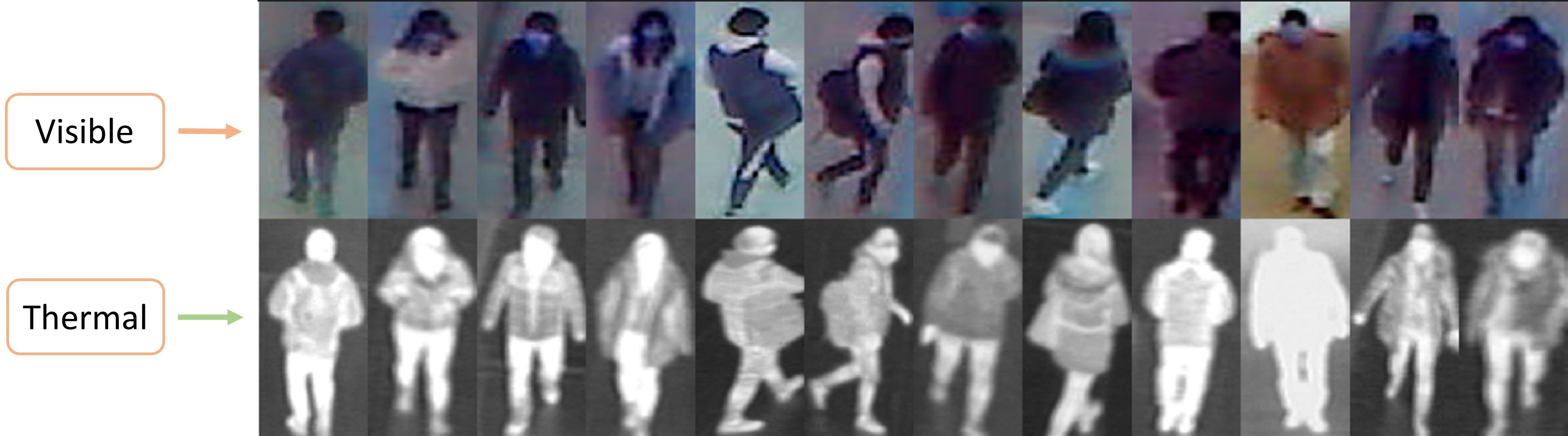}
 }
  \caption{Cross-modality person re-identification. Examples of pairs of persons from the RegDB dataset captured by a dual-camera system (visible and thermal).}
    \label{fig_datasets_examples_regdb}
\end{figure}

\begin{figure}
  \centering
  \subfigure{
  \includegraphics[width=0.97\linewidth]{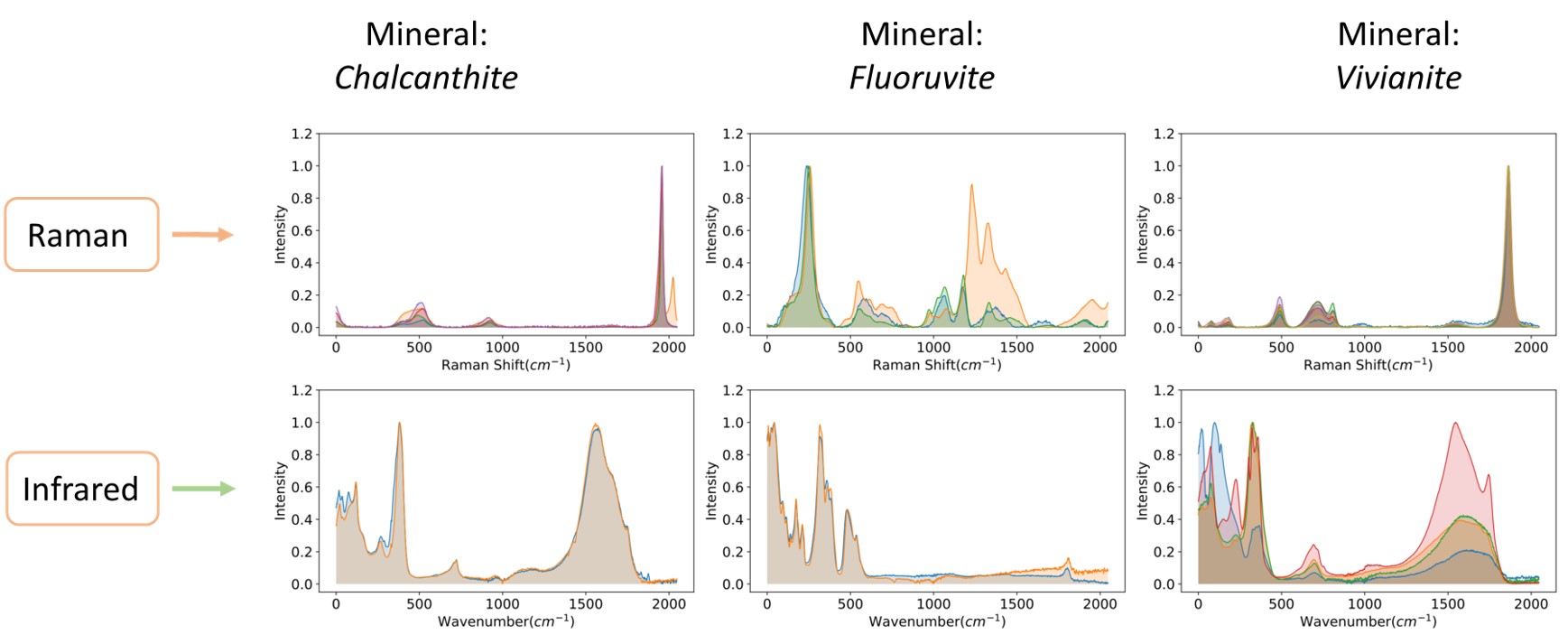}
 }
  \caption{Cross-modality vibrational spectrum matching. Raman spectra (top row) and infrared (bottom row) of the minerals \textit{chalcanthite}, \textit{fluoruvite} and \textit{vivianite}. Unlike HFR or Person ReID, the modality difference between Raman and IR is much more complicated, reflecting different physical processes of two kinds of spectra.}
    \label{fig_datasets_examples_spectrum}
\end{figure}

Additionally, we use a standard triplet loss $L^{\text{triplet}}$ to train the Siamese network. 
\begin{align}
L^{\text{triplet}} &= \sum_{k=1}^{K} \max \Big\{ \Big \|S(E_{1}(M_{1})_{(k)}^{a})  
    - S(E_{2}(M_{2})_{(k)}^{p})\Big \|_{2}^{2}  \\ \nonumber
    &- \Big \|S(E_{1}(M_{1})_{(k)}^{a}) - S(E_{2}(M_{2})_{(k)}^{n})\Big \|_{2}^{2} +  \alpha, 0\Big\} 
\end{align}
where $K$ is the number of the triplets. $\alpha$ is a margin.
Putting all together we get the overall loss:
\begin{equation}
\begin{aligned}
L&=\gamma_{1}(L^{adv}_{11}+L^{adv}_{21})+\gamma_{6}(L^{adv}_{12}+L^{adv}_{22}) \\
 & +\gamma_{2}L^{cross}_{12}+ \gamma_{5}L^{cross}_{21}+\gamma_{3}L^{rec}_{1}+\gamma_{4}L^{rec}_{2}  \\ &+ \gamma_{7}L^{triplet}
 \label{Eq_all_loss_terms}
\end{aligned}
\end{equation}
where $\{ \gamma_{i}, i=1,\cdots, 7 \}$ are positive values to balance all the loss terms. We use a Siamese network and triplet loss framework as an typical example. In practice, any backbone with associated losses for feature extraction can be used.

\paragraph{Inference} In inference, the decoders and discriminators are discarded. Network inputs are passed through the encoder corresponding to their modality and then are fed to the Siamese network for matching. %

\begin{table*}
\centering
\caption{Benchmark datasets used in this study}
\begin{tabular}{lccccc}
\toprule
Dataset   & Applications & Modality & \# Classes &\# Samples &  \# Samples/class\\
    \midrule
 \multirow{2}*{cmRRUFF}  & \multirow{2}*{Mineral Recognition} &   Raman  & \multirow{2}*{360} & 1391   &  $\approx$ 3.86 \\
  & &Infrared & & 622   &  $\approx$ 1.73\\
\addlinespace[1.5ex]
 \multirow{2}*{CUFSF} & \multirow{2}*{Face Recognition} & Photo  & \multirow{2}*{1194} & 1194 &  1 \\
  & &Sketch & & 1194   &  1\\
\addlinespace[1.5ex]
   \multirow{2}*{CASIA NIR-VIS 2.0}  & \multirow{2}*{Face Recognition} & Visible  & \multirow{2}*{725} & $\approx$12.5K  &  1-22  \\ 
    & & Near Infrared  & &  $\approx$5K & 5-50 \\
\addlinespace[1.5ex]
\multirow{2}*{IRIS Thermal/Visible Face Database}  & \multirow{2}*{Face Recognition} & Visible  & \multirow{2}*{31} & 2891  &  47-496\\ 
    & & Thermal  & &  1237  & 1-244 \\
    \addlinespace[1.5ex]
\multirow{2}*{RegDB}  & \multirow{2}*{Person Re-identification} & Visible  & \multirow{2}*{412} & 4120  &  10\\ 
    & & Thermal  & &  4120  & 10 \\
\bottomrule
\end{tabular}
\label{Tab:Dataset}
\end{table*}

 \begin{figure}
  \centering
  \includegraphics[width=0.8\columnwidth]{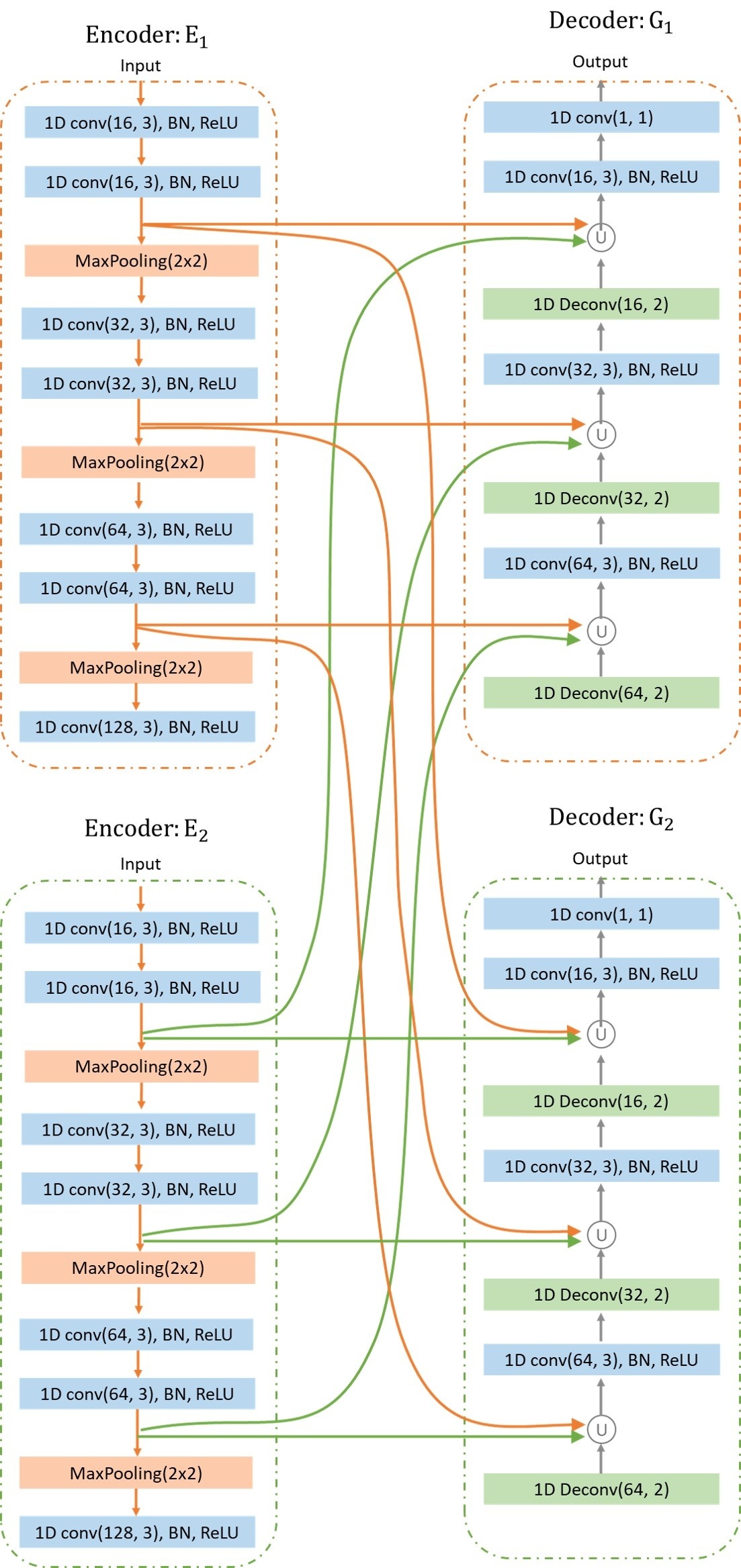}
  \caption{Detailed architecture of cmUNet for cross-modality spectrum matching. \Circled{U} stands for feature concatenation. Note that shortcuts from encoders to decoders of the other modality were added to facilitate the modality transfer.}
    \label{fig_diagram_cmunet_spectrum}
\end{figure}

\begin{table*} %
  \centering %
    \caption{Performance comparison on cmRRUFF for cross-modality vibrational spectrum matching. Style: \textbf{best} and \uline{second best}}\label{tab_result_RRUFF}
   \begin{tabular}{lccccccc}
  \toprule %
   Method  & Modality Transfer &Recall@1 & Recall@3 & Recall@5 & mAP@1 & mAP@3 &mAP@5\\
  \midrule
   Siamese-LeNet & - &  24.08 & 44.48 & 58.04 &24.08 & 32.82 & 35.92\\ 
    {cmGAN~\cite{ijcai2018p94}} & - & {22.01} & {40.14} & {52.82} & {22.02} & {29.88} & {32.86} \\
    {WCNN~\cite{WCNN_TPAMI2019}} & - & {26.34} & {41.88} & {49.86} & {26.34} & {33.14} & {34.94}  \\
   \addlinespace[1ex]
  \multirow{3}*{FCN~\cite{FCN2017}}& Raman$\rightarrow$IR & 26.78 & 47.54 & 59.75 &26.78 & 35.91 & 38.67\\
   &IR$\rightarrow$Raman & 26.95 & 50.16 & 60.17 &26.95 & 37.22 & 41.16\\
    &bidirectional+fusion & \uline{32.14} & \uline{52.29} & \uline{63.63} & \uline{32.14} & \uline{40.86} & \uline{43.40}\\ 
    \addlinespace[1ex]
   \multirow{3}*{CycleGAN~\cite{zhu2017unpaired}}
   & Raman$\rightarrow$IR & 27.08 & 50.20 & 59.64 &27.08 & 37.30 & 39.44\\
  & IR$\rightarrow$Raman & 22.72 & 43.30 & 57.44 &22.72 & 31.66 & 34.86\\
    & bidirectional+fusion& 28.97 & 49.20 & 62.22 &28.97 & 37.69 & 40.63\\
    \addlinespace[1ex]
  \multirow{3}*{IACycleGAN~\cite{fang2020identity}}
   & Raman$\rightarrow$IR & 25.76 & 47.02 & 58.06 &25.76 & 34.70 & 34.72\\
  & IR$\rightarrow$Raman & 23.72 & 43.20 & 55.36 &23.72 & 32.26 & 35.08\\
    & bidirectional+fusion & 25.68 & 48.43 & 59.41 &25.68 & 35.50 & 38.03\\
  \midrule  
  \rowcolor{black!20} 
   MarrNet (ours) & - &  \textbf{36.58} & \textbf{56.41} & \textbf{67.76} &\textbf{36.58} & \textbf{44.75} & \textbf{47.34}\\ 
  \bottomrule%
  \end{tabular}
\end{table*}

\section{Experiments}\label{sec_experiments}
To validate the proposed method, we conducted experiments and analysed our MarrNet along with a number of state-of-the-art approaches on five challenging tasks from three distinct domains, namely vibrational spectroscopy, heterogeneous face recognition and cross-modality person re-identification, as shown in Table~\ref{Tab:Dataset}. The first represents a challenging task of few-shot learning where we found that explicit modality transfer between Raman and infrared spectra is of great difficulty. In contrast, for heterogeneous many cross-modality recognition methods have been suggested with explicit modality transfer or generative models~\cite{1238414,1467376,4624272,zhang2015end,wang2018back,kazemi2018unsupervised,DVGFace_TPAMI2022,fang2020identity}. 

The difference between modalities depends on the application. In CASIA NIR-VIS 2.0 dataset, the modality gap is relatively small, as face shapes are similarly in visible and near infrared images, and the main difference is due to texture. The modality gap in the IRIS Thermal/Visible Face Database is much larger.  CUFSF database includes photos and sketches, which not only differ in shape and texture of faces, but also include cognitive variations introduced by the artists who drew the sketches.

In principle, the modality differences in these datasets may be captured by a statistical model (realized by e.g. neural networks) based on pixel (patch) level correspondence (after straightforward scaling or rotation if needed).

The two modalities in cmRRUFF dataset belong to the vibrational spectroscopy, specifically, Raman and IR. These modalities are linked by complex physical processes with (chemical) band-to-band correspondence\footnote{There is not apparent wavenumber-to-wavenumber correspondence between Raman and infrared spectra.}. It is much more difficult to learn the relationship between these two modalities, compared to those in the face datasets. These five datasets serve as quality benchmarks for evaluating the generalizability of tested methods.

\subsection{Cross-Modality Vibrational Spectrum Matching}\label{subsec_cmvsm}

\subsubsection{Dataset and Evaluation Protocol} 
For cross-modality vibrational spectrum recognition, we created ``cmRRUFF'' dataset out of the RRUFF database~\cite{RRUFFdataset} that contains both Raman and infrared spectra of 360 minerals. Each sample in cmRRUFF has at least one Raman and one infrared spectra, and there are 1391 Raman and 622 IR spectra in total. Examples of pairs of Raman and infrared spectra are plotted in Fig.~\ref{fig_datasets_examples_spectrum}. The x-axis of Raman and infrared spectra is Raman shift and wavenumber respectively, and the y-axis is the intensity. For each mineral, multiple spectra of Raman and infrared have plotted to illustrate the variations within and cross modalities. 

We randomly partitioned cmRRUFF into training, validation and test sets, using 300 samples including both Raman and IR spectra for training, 30 for validation and 30 for testing. We repeated the experiments with 5 random partitions and reported the averaged results. 

We employed two evaluation metrics, \textbf{Top-$k$ Accuracy} (Recall@k) and \textbf{Mean Average Precision $k$} (mAP@k), which are widely used in image retrieval to evaluate the predictive performance of models. Top-$k$ Accuracy measures the number of successful matches where queries are among the closest $k$ neighbors. Mean Average Precision $k$ calculates the averaged precision of the $k$ closest neighbors\cite{Soren2021_mAP}.

\subsubsection{Network Architecture and Training Protocol}

Our network consists of a cmUNet including two encoders and two decoders, two discriminators and a Siamese network for feature extraction. The encoders in cmUNet contain three blocks, each comprising two one dimensional convolutional layers with batch normalization and ReLU activation followed by a max-pooling layer. At each downsampling step we double the number of feature channels. The last layer of the encoders is a convolutional layer with batch normalization(BN) and ReLU activation that outputs a 128-dimension feature vector. The decoders contain 3 block comprising transposed convolution layer followed by a convolutional layer of sizes matching the encoder. The last layer in the decoders is 1-by-1 convolutional layer. The bridge connections go from each encoder to both decoders (see Fig.~\ref{fig_diagram_cmunet_spectrum} for the detailed architecture of the cmUNet).

The discriminators have two blocks that match the architecture of the encoders, followed by a linear layer with two outputs. The Siamese network consists of three blocks each with two convolutional layers with BN and ReLU activation followed by max pooling. The number of kernels of each block is 16, 32 and 64 respectively. The size of the kernel is 3. The last layer is linear. The constants for balancing the loss terms are set to $\gamma_{i \in \{1,6\}}=0.01, \gamma_{i \in \{2,3,4,5\}}=0.001, \gamma_{7}=1$. For the triplet loss, $K=5, \alpha=1.0$. Data augmentation for vibrational spectroscopy has been applied to all the compared methods.

For cross-modality spectrum matching, all the modules $\{E_{1}, E_{2}, G_{1}, G_{2}, D_{1}, D_{2}, S \}$ which contain trainable weights were trained from random weights by end-to-end optimization using Adam. The maximum number of epochs was set to 600. The initial learning rates were set to 1e-4 for $\{E_{1}, E_{2}, G_{1}, G_{2}, S \}$, and 1e-3 for $D_{1}, D_{2}$, then decreased by 25\% every 10 epochs. 

\subsubsection{Results and Analysis}

We evaluated our MarrNet and compared with state-of-the-art methods including 
cmGAN\cite{ijcai2018p94}, WCNN\cite{WCNN_TPAMI2019}, IACycleGAN~\cite{fang2020identity}, CycleGAN~\cite{zhu2017unpaired,fang2020identity} and FCN~\cite{zhang2015end}. Among these methods, cmGAN uses a modality-classifier (discriminator) combined with cross-modality triplet loss to learn modality-invariant representations. WCNN makes use of the Wasserstein loss to reduce the modality/domain differences. The other three methods are equipped with explicit image-to-image modality transfer to bridge the modality gap, where FCN was included as a baseline. We note that methods such as DVG-Face\cite{DVGFace_TPAMI2022} that show superior performance in heterogeneous face recognition are not applicable here due to their need for a large auxiliary dataset for training that does not exist in this experiment.

The results reported in Table \ref{tab_result_RRUFF} show that the proposed MarrNet significantly outperforms all  other methods in all evaluation metrics. It is worth noting that the previous methods performed poorly for this application. In particular, the performance of IACycleGAN which was specially-designed for cross-modality matching was even inferior to CycleGAN and the baseline FCN. We hypothesize that the inferior performance can be explain by the fact that they heavily rely on the explicit transfer of modalities. This is, unfortunately, a very challenging task in cross-modality spectrum matching, since the models need to infer the physical processes (or inverses) of Raman and IR spectroscopy. When the transferred samples are of bad quality, they make the matching harder. Especially here only a limited amount of data is available, both cmGAN and WCNN performed poorly. Alternatively, our method realized an implicit modality transfer(cmUNet), which served as a regularizer to the learning of the modality-agnostic representations. Our cmUNet module addressed the problem of limited labeled data by a self-supervised training and thus achieved significantly better results.

\begin{table} %
    \centering %
    \caption{Performance comparison on CUFSF for heterogeneous face recognition. Style: \textbf{best} and \uline{second best}}
    \begin{tabular}{lcc}
    \toprule %
     Method & Modality Transfer & Recall@1 \\
    \midrule%
    \multicolumn{3}{c}{\colorbox {cyan!20}{\textbf{Protocol A:}}} \\
    \addlinespace[1.5ex]
    \multirow{1}*{Siamese-VGGFace}  & - &57.94 \\
    {cmGAN\cite{ijcai2018p94}} & - &{61.94} \\
    {WCNN\cite{WCNN_TPAMI2019}} & - &{58.00} \\
    \addlinespace[1ex]    
    \multirow{3}*{FCN~\cite{FCN2017}} 
    &photo$\rightarrow$sketch & 24.62  \\
    &sketch$\rightarrow$photo & 28.31  \\
    &bidirectional+fusion & 40.44  \\    
    \addlinespace[1ex]
    \multirow{3}*{Conditional-GAN\cite{isola2017image}}
     &photo$\rightarrow$sketch  &41.21  \\
     &sketch$\rightarrow$photo  &28.07  \\
    &bidirectional+fusion &57.94 \\
    \addlinespace[1ex]
    \multirow{3}*{CycleGAN\cite{zhu2017unpaired}} 
    &photo$\rightarrow$sketch & 68.75  \\
    &sketch$\rightarrow$photo & 59.00  \\
    &bidirectional+fusion & 73.52  \\
    \addlinespace[1ex]
    \multirow{3}*{IACycleGAN\cite{fang2020identity}}
      &photo$\rightarrow$sketch &70.87  \\
     &sketch$\rightarrow$photo &64.94  \\
     &bidirectional+fusion & \uline{81.36}  \\
    \midrule
    \rowcolor{black!20} 
    \textbf{MarrNet (ours)} & - & \textbf{82.52} \\
    \midrule
    \multicolumn{3}{c}{\colorbox {cyan!20}{\textbf{Protocol B:}}} \\
    \addlinespace[1.5ex]
    \multirow{2}*{PS$^{2}$-MAN~\cite{wang2018high}} &photo$\rightarrow$sketch &51  \\
     & sketch$\rightarrow$photo & 47  \\
     \addlinespace[1ex]
     Divergent-Model~\cite{NAGPAL2021PR}& - & 67.3 \\
    \addlinespace[1ex]
    \multirow{3}*{CycleGAN~\cite{zhu2017unpaired}} &photo$\rightarrow$sketch &87.54\\
     & sketch$\rightarrow$photo & 86.53\\
     &bidirectional+fusion &93.84\\
    \addlinespace[1ex]
    \multirow{3}*{IACycleGAN~\cite{fang2020identity}} &photo$\rightarrow$sketch &89.56 \\
     & sketch$\rightarrow$photo & 90.91\\
     &bidirectional+fusion & \uline{95.23}\\
    \midrule
    \rowcolor{black!20} 
    \textbf{MarrNet (ours)} & - & \textbf{97.31} \\
    \bottomrule
    \end{tabular}
   \label{tab_result_CUFSF}     
\end{table}

\subsection{Heterogeneous Face Recognition}\label{subsec_hfr}

\subsubsection{Datasets and Evaluation Protocols}\label{hfr_subsubsec_dataset}

We demonstrate the applicability of our method for heterogeneous face recognition (HFR) which involves matching identities across images of different modalities. We chose three widely-used public datasets for photo-sketch, visible-infrared and visible-thermal face recognition respectively.

\textit{CUFSF}: The CUFSF dataset includes photos and corresponding hand-drawn sketches of 1194 human subjects. Following the protocol from the previous work e.g.~\cite{fang2020identity}, 994 individuals were randomly selected as the test set, 200 as the training set and 50 as the validation set. All photos and sketches were normalized and cropped to $256\times256$ pixels. We will refer to this protocol as Protocol A. Note that a different protocol has also been used by recent methods e.g. \cite{wang2018high,NAGPAL2021PR}, we also evaluated the compared methods under this protocol which is referred to as Protocol B. According to this protocol, 297, instead of 944, individuals were selected to form the test set and the rest were used for training. In general this represents an easier task to solve. We have also compared the related methods under this protocol.

\begin{table} 
    \centering 
    \caption{Performance comparison on CASIA NIR-VIS 2.0 for heterogeneous face recognition. Style: \textbf{best}}
    \begin{tabular}{llcc}
    \toprule 
     Method & Recall@1 & VR@FAR=0.1\% \\
    \midrule
    \addlinespace[1.5ex]
    IDNet\cite{IDNet2016}  & 87.1$\pm$0.9 &74.5 \\
    HFR-CNN\cite{HFRCNN2016}  & 85.9$\pm$0.9 &78.0 \\
    Hallucination\cite{Lezama2016NotAO}  & 89.6$\pm$0.9 & - \\
    \addlinespace[1ex]   
    TRIVET\cite{TRIVET2016} & 95.7$\pm$0.5 & 91.0$\pm$1.3\\
    DLFace\cite{Peng2019DLFaceDL}  & 98.7 & - \\
    WCNN\cite{WCNN_TPAMI2019} & 98.7$\pm$0.3 & 98.4$\pm$0.4\\
    PACH\cite{PACH2020} & 98.9$\pm$0.2 & 98.3$\pm$0.2\\
    \addlinespace[1ex]  
    RCN\cite{RCN2019} & 99.3$\pm$0.2 & 98.7$\pm$0.2\\
    MC-CNN\cite{MCCNN} & 99.4$\pm$0.1 & 99.3$\pm$0.1\\
    DVR\cite{DVR2018} & 99.7$\pm$0.1 & 99.6$\pm$0.3  \\
    \textbf{DVG-Face}\cite{DVGFace_TPAMI2022} & \textbf{99.9}$\pm$0.1 & \textbf{99.9}$\pm$0.0\\
    \midrule
    \rowcolor{black!20} 
    MarrNet (ours) & 99.6$\pm$0.2 & 99.4$\pm$0.2\\
    \bottomrule
    \end{tabular}
   \label{tab_result_CASIA}     
\end{table}

\begin{table} 
    \centering
    \caption{Performance comparison on IRIS Thermal/Visible Face Database for heterogeneous face recognition. Style: \textbf{best} and \uline{second best}}
    \begin{tabular}{lcc}
    \toprule 
     Method & Modality Transfer & Recall@1 \\
    \midrule
    \addlinespace[1.5ex]
    \multirow{1}*{Siamese-VGGFace}  & - &69.36 \\
    {cmGAN\cite{ijcai2018p94}} & - &{74.20} \\
    {WCNN\cite{WCNN_TPAMI2019}} & - &{\uline{85.44}} \\
    \addlinespace[1ex]    
    \multirow{3}*{FCN~\cite{FCN2017}} 
    &visible$\rightarrow$thermal & 25.75  \\
    &thermal$\rightarrow$visible & 39.92  \\
    &bidirectional+fusion & 43.11  \\    
    \addlinespace[1ex]
    \multirow{3}*{Conditional-GAN\cite{isola2017image}}
     &visible$\rightarrow$thermal  &62.98  \\
     &thermal$\rightarrow$visible  &58.85  \\
    &bidirectional+fusion &76.77 \\
    \addlinespace[1ex]
    \multirow{3}*{CycleGAN\cite{zhu2017unpaired}} 
    &visible$\rightarrow$thermal & 64.44  \\
    &thermal$\rightarrow$visible & 59.45  \\
    &bidirectional+fusion & 76.15  \\
    \addlinespace[1ex]
    \multirow{3}*{IACycleGAN\cite{fang2020identity}}
      &visible$\rightarrow$thermal &70.87  \\
     &thermal$\rightarrow$visible &64.94  \\
     &bidirectional+fusion & 72.89  \\
    \midrule
    \rowcolor{black!20} 
    \textbf{MarrNet (ours)} & - & \textbf{88.40} \\
    \bottomrule
    \end{tabular}
   \label{tab_result_IRIS}     
\end{table}

\begin{table*}[h]
  \centering %
  \caption{Performance comparison on the RegDB dataset for cross-modality person ReID. Style: \textbf{best} and \uline{second best}}
   \begin{tabular}{lcccccccc}
  \toprule %
  \multirow{2}*{Method} & \multicolumn{4}{c}{Visible to Thermal} & \multicolumn{4}{c}{Thermal to Visible}\\
     &R-1 & R-10 & R-20 & MAP & R-1 & R-10 & R-20 & MAP \\
  \midrule%
  NFS~\cite{chen2021neural}     & 80.5 & 91.6 & 95.1 & 72.1 & 78.0 & 90.5 & 93.6 & 69.8\\
  CM-NAS~\cite{fu2021cm}        & 82.8 & 95.1 & 97.7 & 79.3 & 81.7 & 94.1 & 96.9 & 77.6\\
  MPANet~\cite{wu2021discover}  & 82.8 & -    & -    & 80.7 & 83.7 & - & - & 80.9\\
  SMCL~\cite{wei2021syncretic}  & 83.9 & -    & -    & 79.8 & 83.1 & - & - & 78.6\\
  DART~\cite{yang2022learning}  & 83.6 & -    & -    & 75.7 & 82.0 & - & - & 73.8\\
  CAJ~\cite{ye2021channel}      & 85.0 & 95.5 & 97.5 & 79.1 & 84.8 & 95.3 & 97.5 & 77.8\\
  MAUM~\cite{liu2022learning}   & 87.9 & -    & -    & \textbf{85.1} & 87.0 & - &  - & \textbf{84.3}  \\
  FMCNet~\cite{zhang2022fmcnet} & 89.1 & -    & -    & 84.4 & 88.4 & - & - & \uline{83.9}\\
  MMN~\cite{zhang2021towards}   &\uline{91.6} & 97.7 & \uline{98.9} & 84.1 & 87.5 & 96.0 & 98.1 & 80.5\\ 

  PMT~\cite{PMT2023}   & 84.8 & -    & -    & 76.6 & 84.2 & - & - & 75.1\\        
  DPIS~\cite{ShiDualPI}   & 85.6 & -    & -    & 76.7 & 81.4 & - & - & 74.1 \\        
  DEEN~\cite{zhang2023diverse}  & 91.1 & \uline{97.8}  & \uline{98.9} & \textbf{85.1} & \uline{89.5} & \uline{96.8} & \uline{98.4} & 83.4\\
  \midrule%
  \rowcolor{black!20} 
MarrNet(ours) & \textbf{92.6} & \textbf{98.2} & \textbf{99.3} & \uline{84.7} & \textbf{91.0} & \textbf{97.9} & \textbf{99.0} & 83.3\\
  \bottomrule%
  \end{tabular}
\label{tab_RegDB_results}
\end{table*}

\textit{CASIA NIR-VIS 2.0 Face Database}: This is the largest publicly-available and challenging dataset of identifying faces across visible and near infrared modalities\cite{CASIANIRVIS2013}. Besides variations induced by the modality gap, images in this dataset contain other variations including lighting, pose, expression and presence or absence of glasses. This dataset contains 725 subjects, each of which has 1-22 visible and 5-50 near infrared images. We followed the standard evaluation protocol in ``View 2'' to evaluate the tested methods\cite{CASIANIRVIS2013,WCNN_TPAMI2019,DVGFace_TPAMI2022}.

\textit{IRIS Thermal/Visible Face Database}: This dataset includes  visible and thermal facial images with variation in pose, expression and glasses attribute. The dataset contains 31 subjects, each of which has 47-496 visible and 1-244 thermal face images. We randomly chose all the photos of 10\% of all the subjects as test set, and the rest used for training and validation. We reported averaged results on five random partitions.

\subsubsection{Network Architecture and Training Protocol}

Due to the small size of the CUFSF dataset, previous methods have used a pretrained backbone, namely VGG16, to reduce overfitting which has been adopted in our experiment for a fair comparison.

We partitioned the VGG16 into two parts layer-wise: we replicated the upper layers of VGG16 and used them as the encoders $E_{1}$ and $E_{2}$. The decoders $G_{1}, G_{2}$ complemented the architecture to form a cross-modality UNet. We used the bottom layers of the VGG16 for the Siamese network (as it inputs feature vectors). During training, the weights of the Siamese network were frozen for the first 300 epochs and then fine-tuned along with the rest of the network.

The architectures of the discriminators for this applications are similar to those for spectrum matching, except that here convolutional layers are two-dimensional and have double amount of trainable kernels. 
The Siamese network consists of three blocks each with three convolutional layers with ReLU activation followed by max pooling. The last layer is linear. The constants for balancing the loss terms are set to $\gamma_{i \in \{1,6\}}=0.01, \gamma_{i \in \{2,3,4,5\}}=0.001, \gamma_{7}=1$. For the triplelet loss, $K=4, \alpha=0.3$.

All the modules $\{E_{1}, E_{2}, G_{1}, G_{2}, D_{1}, D_{2}, S \}$ which contain trainable weights were trained from random weights by end-to-end optimization using Adam. The maximum number of epochs was set to 600. The initial learning rates were set to 1e-4 for $\{E_{1}, E_{2}, G_{1}, G_{2}, S \}$, and 1e-3 for $D_{1}, D_{2}$, then decreased by 25\% every 40 epochs. 

For experiments on IRIS Thermal/Visible Face Database, the same network architecture and training protocol has been employed, while for CASIA NIR-VIS 2.0, we adopted the backbone used in \cite{DVGFace_TPAMI2022}. The details are omitted here due to page limit.

\begin{table*}[h] %
  \centering %
  \caption{Ablation experiments on the CUFSF dataset}

   \begin{tabular}{lccccccc}
  \toprule %
   Method  &Recall@1 & Recall@3 & Recall@5 & MAP@1 & MAP@3 &MAP@5\\
  \midrule%
    $L^{\text{triplet}} $  & 80.49 & 91.89 & 94.58 & 80.49 & 85.66 & 86.30\\ 

    $L^{\text{triplet}} + L^{\text{adv}} + L^{\text{rec}}$  & 74.41 & 87.88 & 91.74 & 74.41 & 80.46 & 81.35\\
    $L^{\text{triplet}} + L^{\text{adv}} + L^{\text{cross}}$& 81.02 & 91.36  & 94.19  & 81.02  & 85.69  & 86.34 \\
    $L^{\text{triplet}} + L^{\text{rec}}+ L^{\text{cross}}$ & 79.64 & 91.76 & 94.45 & 79.64 & 85.13 & 85.74\\    
    $L^{\text{triplet}} + L^{\text{rec}}+ L^{\text{cross}}+L^{\text{adv}}$ (w/o shortcuts)& 80.30 & 91.63 & 94.83 & 80.30 & 85.41 &86.16\\
    \addlinespace[1ex]
    \rowcolor{black!20} 
    $L^{\text{triplet}} + L^{\text{rec}}+ L^{\text{cross}}+L^{\text{adv}}$ & 82.52 & 92.88 & 95.66 & 82.52 & 87.21 & 87.46\\ 
  \bottomrule%
  \end{tabular}
\label{tab_ablation_CUFSF}
\end{table*}

\begin{table*}[h] %
  \centering %
  \caption{Model complexity and Run-time of MarrNet and other compared methods for cross-modality spectrum matching.}

  \setlength{\tabcolsep}{1mm}{
   \begin{tabular}{l|cccc|cc}
  \toprule %
  \multirow{2}*{Method}  & \multicolumn{4}{c|}{Training} & \multicolumn{2}{c}{Inference}\\  
     &\makecell[c]{Model Size\\(MB)} & \makecell[c]{GPU Memory\\(GB)} &\makecell[c]{Forward Pass\\(ms))} & \makecell[c]{Backward Pass\\(ms)} & \makecell[c]{GPU Memory\\(GB)} &\makecell[c]{Forward Pass\\(ms)}\\
  \midrule%
    Siamese-LeNet                      & 4.30  & 0.82 & 1.15  & 7.70    & 0.82 & 2.15 \\
    FCN\cite{FCN2017}                  & 29.20 & 2.73 & 6.14  & 10.94   & 2.73 & 3.12 \\
    CycleGAN\cite{zhu2017unpaired}     & 35.93 & 2.22 & 22.53 & 43.21   & 1.89 & 8.24 \\
    IACycleGAN\cite{fang2020identity}  & 54.00 & 1.69 & 23.92 & 50.78   & 1.33 & 7.82 \\
    {cmGAN\cite{ijcai2018p94}}   & {8.20} & {2.10} & {5.64}  & {18.56}  &  {0.82} & {2.15} \\
{WCNN\cite{WCNN_TPAMI2019}}  & {18.42} & {2.15} & {6.12}  & {20.16}  &  {0.90} & {2.18} \\
    \midrule
    \rowcolor{black!20} 
    MarrNet(Ours)                      & 5.52  & 1.73 & 8.63  & 30.53   & 1.05 & 1.52 \\
  \bottomrule%
  \end{tabular}}
\label{model_complexity_spectrum}
\end{table*}

\subsubsection{Results and Analysis} 

For CUFSF, we evaluated the proposed MarrNet along with a number of state-of-the-art methods including cmGAN\cite{ijcai2018p94}, WCNN\cite{WCNN_TPAMI2019}, IACycleGAN\cite{fang2020identity}, CycleGAN~\cite{zhu2017unpaired} and conditional-GAN~\cite{isola2017image}, PS$^{2}$-MAN~\cite{wang2018high} which use explicit modality transfer, and Divergent model~\cite{NAGPAL2021PR} that applies implicit modality transfer and learns cross-domain(modality) representations. We also included the results of matching feature vectors extracted by the VGGFace as a baseline, denoted as Siamese-VGGFace. 

For both protocols, five random partitions were generated for evaluation and the averaged results were reported in Table~\ref{tab_result_CUFSF}. The results of PS$^{2}$-MAN and Divergent model were taken from corresponding literatures. For comparison, we have also run CycleGAN and IACycleGAN under this protocol and reported the results. It can be seen that under both protocols, our method achieved substantially better performance than existing methods. Among the existing methods, Divergent model performed unsatisfactorily. This might be owing to its intrinsic shortcomings where it learns a simple linear transform from inputs to representations shared by two modalities. In contrast, instead of a shared linear transform our MarrNet used two separate encoders to encode two different modalities.

For CASIA NIR-VIS 2.0, the standard evaluation protocol in View 2 has been used to evaluate our MarrNet and existing  methods\cite{CASIANIRVIS2013,WCNN_TPAMI2019,DVGFace_TPAMI2022}. Results are reported in Table~\ref{tab_result_CASIA} where MarrNet has achieved competitive performance comparable to existing state-of-the-art methods. 

Results on IRIS Thermal/Visible Face Database are reported in Table~\ref{tab_result_IRIS} showing that MarrNet achieves the superior performance compared to other tested methods. It is interesting to see that WCNN outperformed all the methods with explicit modality transfer and achieved second best result. As shown in Fig.~\ref{fig_datasets_examples_face}, converting between thermal and visible images appeared to be more difficult than the other two tasks, which could explain the poor performance of methods with explicit modality transfer.

\begin{table}[h]
  \centering 
  \caption{Analysis of Parameters weighting the loss terms in Eq.~(\ref{Eq_all_loss_terms}) on the CUFSF dataset}
   \begin{tabular}{lccccccc}
  \toprule 
  $\gamma_{1}=\gamma_{6}$ & $\gamma_{2}=\gamma_{3}=\gamma_{4}=\gamma_{5}$ &$\gamma_{7}$ & 
  \multirow{2}*{Recall@1}\\
   ($\times$E-2) & ($\times$E-3)  &  & \\
  \midrule
    1  & 0.25 & 1 & 78.50 \\
    1  & 0.5 & 1 & 79.47 \\
    1  & 0.75 & 1 & 80.51 \\
    1  & 1 & 1 & 82.52 \\
    1  & 1.25 & 1 & 80.17 \\
    1  & 1.5 & 1 & 81.73 \\
    1  & 1.75 & 1 & 78.62 \\
    1  & 2 & 1 & 76.25 \\
    2  & 1 & 1 & 76.56 \\
  \bottomrule
  \end{tabular}
\label{tab_parameters_CUFSF}
\end{table}

\subsection{Cross-Modality Person Re-identification}

\subsubsection{Dataset and Protocols}
A standard benchmark dataset \textit{RegDB} for visible-thermal person re-identification has been used in our study. A dual camera system where visible and thermal cameras were mounted together to collect pairs of images\cite{wu2017rgb}. RegDB includes 412 identities, each of which contains 10 visible and 10 thermal images. Following the standard protocol, 50\% of the samples were used for training and the rest for testing. Two evaluation modes have been used, \textit{visible to thermal} where matching a visible image against a gallery of thermal images, and  \textit{thermal to visible} where matching a thermal image against a gallery of visible images. Examples of pairs from RegDB are shown in Fig.~\ref{fig_datasets_examples_regdb}.

\subsubsection{Results and Analysis} 

Similar to previous experiments, we adopted the model in \cite{zhang2023diverse} as the backbone which is a state-of-the-art method to learn embeddings for visible-infrared person ReID. Results were reported in Table~\ref{tab_RegDB_results}. Under both evaluation protocols MarrNet achieved superior performance compared to existing methods. Particularly, it improved the Rank-1 accuracy by 1.5\% over DEEN~\cite{zhang2023diverse} through regularizing the learned intermediate features only and without the need of any auxiliary information, demonstrating the merit of modality-agnostic representation learning using the proposed neural module cmUNet.

\begin{figure}
  \centering
  \includegraphics[width=\columnwidth]{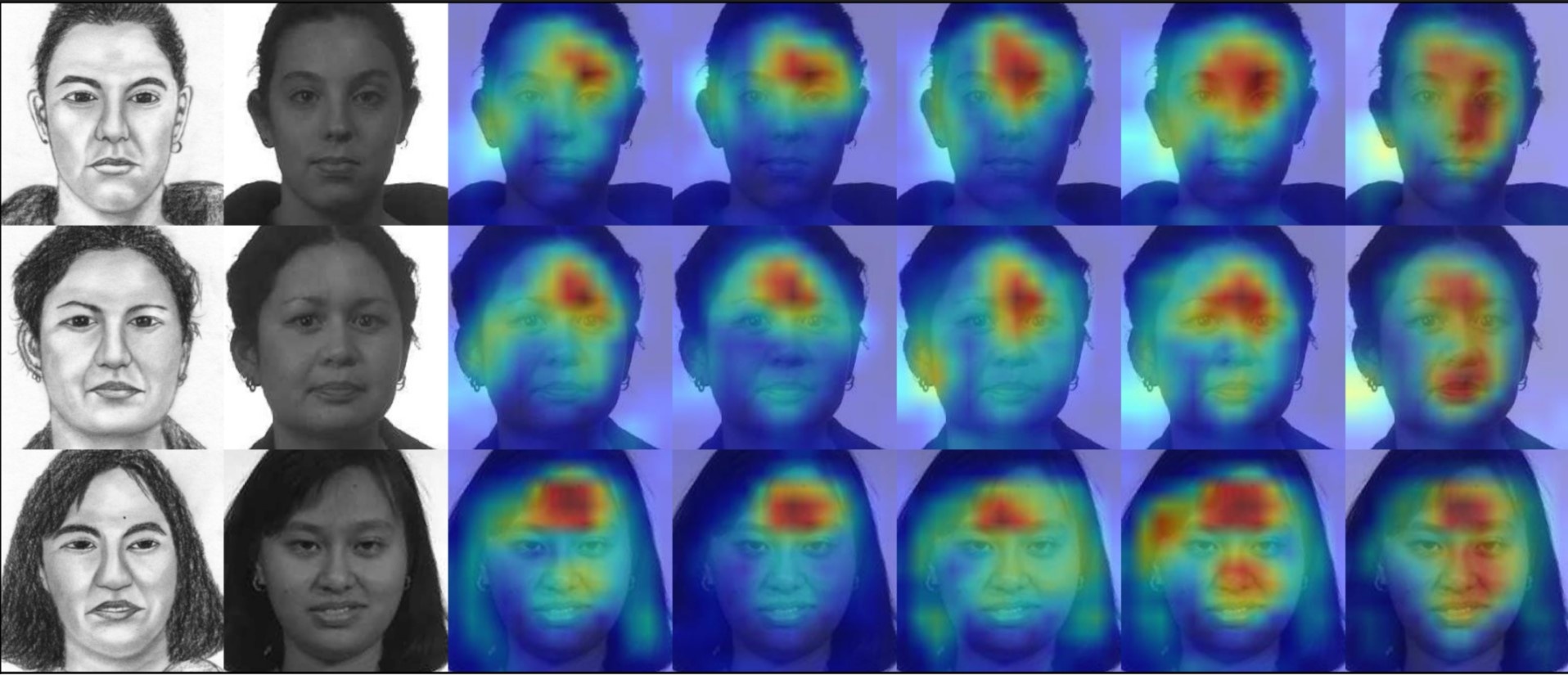}
  \caption{Ablation study of visualizing the saliency maps of MarrNet with different combination of the loss terms. The first two columns are pairs of sketch and photo. The third to seven columns correspond to saliency maps of MarrNet with only $L^{\text{triplet}}$, $L^{\text{triplet}} + L^{\text{rec}}$, $L^{\text{triplet}} + L^{\text{cross}}$, $L^{\text{triplet}} + L^{\text{rec}}+ L^{\text{cross}}$ and $L^{\text{triplet}} + L^{\text{rec}}+ L^{\text{cross}}+L^{\text{adv}}$ respectively.}
    \label{fig_saliencymap_ablation}
\end{figure}

\subsection{Ablation Studies and Parameters Analysis}

\subsubsection{Ablation Studies}
We performed ablation study on CUFSF and reported the results in Table~\ref{tab_ablation_CUFSF}. It is evident that all the components of the proposed MarrNet contributed to the performance improvement on both datasets. In particular, the bridge connections between encoders and decoders in cmUNet are of great importance. Without bridge connections, the performance dropped significantly.

It can also be observed that all the three losses $L^{\textit{rec}}, L^{\textit{cross}}, L^{\textit{adv}}$ functioned best when they were used simultaneously. In fact, when using two of the losses, the performance were comparable or even worse than using only the triplet loss. 
Particularly in Table~\ref{tab_ablation_CUFSF}, the accuracy trained with the losses excluding the cross-transform loss term dropped significantly compared to the baseline. We reckon this happened as the learned representations inherited from the pretrained backbone were damaged, yet still interfered by the modality-related information due to the lack of the cross-modality transform loss term.

\subsubsection{Parameters Analysis}
We analyzed the seven loss terms in Eq.~(\ref{Eq_all_loss_terms}) which balance the learning of the model and reported the results in Table~\ref{tab_parameters_CUFSF}. Note that to make the analysis feasible, similar parameters have been set to same values for the sake of efficiency.  Moreover, to have an intuitive understanding of how the loss terms contribute to the learning, we visualized the saliency maps of the proposed MarrNet with different combination of the loss terms. It can be seen when more loss terms were taken into action, MarrNet gradually focused on the true discriminant regions.

\subsection{Model Complexity and Run-time}
We calculated the model complexity and run-time of our MarrNet along with typical existing methods on a server with a Nvidia RTX 3090 GPU, Intel Xeon(R) Silver 4210R CPU and 128G RAM. Results can be found in Table~\ref{model_complexity_spectrum}. 
For cross-modality spectrum matching, the model size of MarrNet is only 20\% of the second best model FCN. For inference, all the compared methods were comparably efficient.

\begin{figure}
  \centering
  \includegraphics[width=\columnwidth]{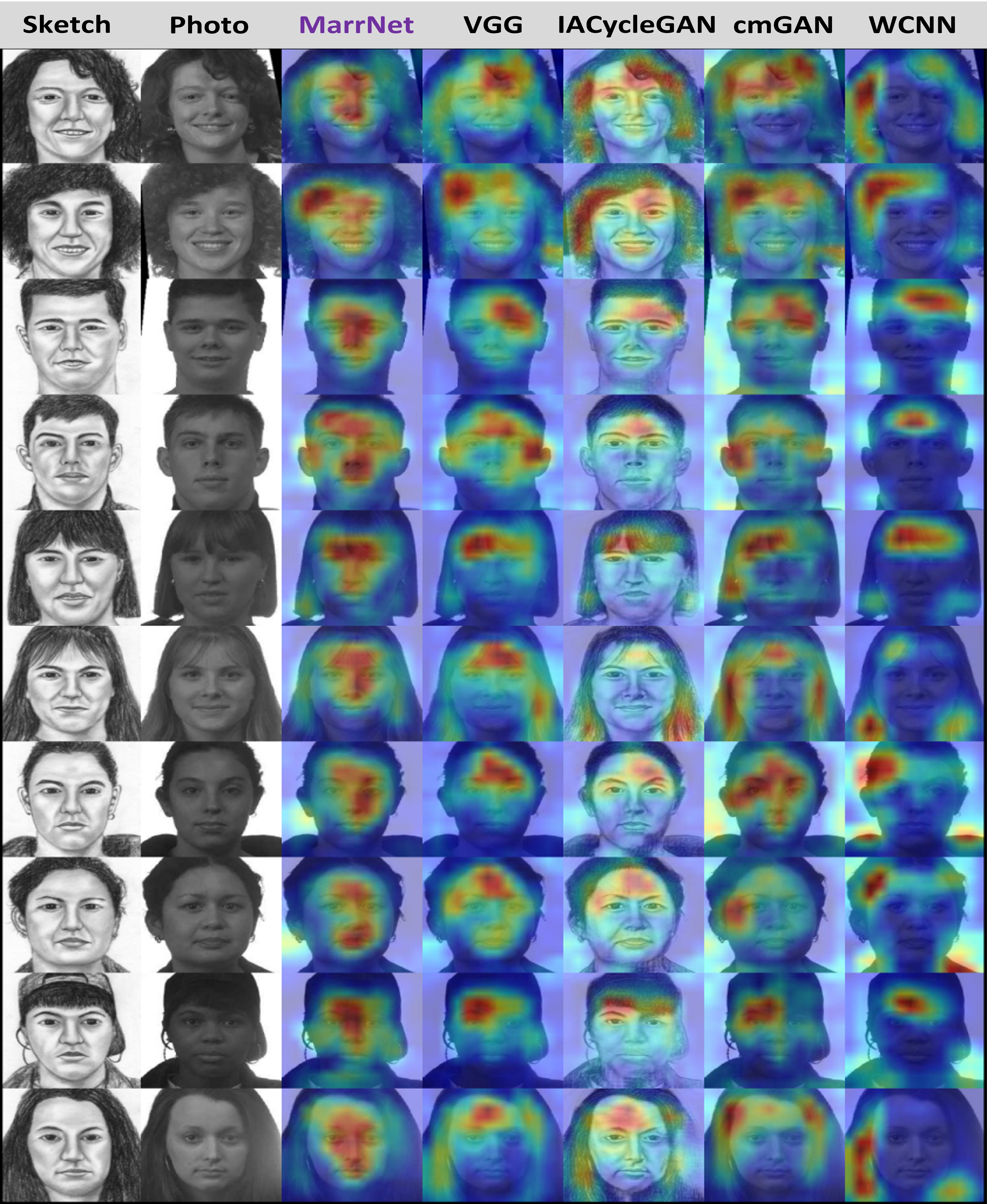}
  \caption{Saliency maps of the compared methods on more examples. The first two columns are the sketches and corresponding photos. For saliency maps, from left to right, correspond to the method of MarrNet(ours), VGG, IACycleGAN, cmGAN and WCNN respectively.}
    \label{fig_saliencymap_face_more}
\end{figure}

\section{Mind the gap: bridging the modality gap matters}
As shown in section \ref{sec_experiments}, the proposed MarrNet achieved superior matching accuracy, which is no doubt an important, but not the sole, criterion to evaluate the learned representations. To further investigate the difference of representations learned by different cross-modality matching methods, especially how these representations are affected by the modality gap, we examined the saliency maps\footnote{We employed a state-of-the-art tool~\cite{stylianou2019visualizing} which is designed to visualize neural attention of deep networks for similarity learning.} of the compared methods at the final representation layer prior to the classification and demonstrated that the proposed MarrNet focused more on the facial regions while the existing methods utilized more information from non-facial regions, as illustrated in Fig.~\ref{fig_saliencymap_face_more}. 
It can be seen that the modality gap caused by the cognitive drawing process is clearly larger in the facial region than that in the non-facial region including hairs and forehead. Humans are especially good at recognising faces across this (cognitive) modality gap, while for learning machines it is rather challenging. It can be seen that except for our MarrNet, all the other methods tended to use non-facial features such as forehead and hairs where modality gap is significantly smaller. This indicated that MarrNet had bridged the modality gap well which explained the superiority of our method. More examples can be found in Fig.~\ref{fig_saliencymap_face_more}.

Inspired by this observation, we proposed a ``thin-ice hypothesis" to explain the effect of modality gap on the representation learning and linked it to robustness to occlusions of models. More specifically, we proposed that robustness to disguises and occlusions can be a quantitative indicator of whether a method can cope with modality gap well. Besides, methods being robust to disguises and occlusions could also be useful in the law enforcement applications where suspects or persons of interest may alter their non-facial features especially hairstyles or wear masks to deceive law enforcement agencies and the public.

\subsection{``Thin-Ice Hypothesis"}
How was MarrNet able to focus on the facial features more than other methods? By taking a closer look at the photo-sketch pairs shown in Fig.~\ref{fig_datasets_examples_face} and \ref{fig_face_disguise_examples}, we can observe that the non-facial regions at two modalities are visually more similar to one another than the facial regions. In other words, the modality gap is smaller in the non-facial regions than the facial ones.
This inspired us to propose the following hypothesis: 
\vspace{0.1cm}

``\textbf{Thin-Ice Hypothesis}": \textit{If a method does not bridge the modality gap well, it's attention may be biased towards regions where the gap is smaller, which then leads to wrongly-focused discriminant representations and ultimately poor generalization.}     

\vspace{0.1cm}
We argue that this is primarily due to the competing nature of bridging the modality gap and learning discriminant representations. When the modality gap is complex or large, learning discriminant representations may dominate the training process and cause premature ending of the process of bridging the modality gap.
We named this \textit{``thin-ice hypothesis"} after the famous quote: ``Ice cracks at the thinnest  point" as an analogy to ``the learning process of discriminant representations may be wrongly-focused on the regions with smallest modality gap (thinnest point)".

Therefore, for methods with poor quality modality transfer or incompetence of bridging modality gaps, the true discriminant region could be largely or completely ignored, see Fig.~\ref{fig_saliencymap_face_more}. As a result, the subsequent discriminant feature learning process could be severely disrupted and produce wrongly-focused representations. Ultimately, the matching in the end could face catastrophic failure. An example is the conditional-GAN shown in Fig.~\ref{fig_match_acc_with_disguises}. More details are shown in the next subsection. In contrast, MarrNet bridges the modality gap well by explicitly learning modality-agnostic representations, and is able to focus on the true discriminant regions.

To validate this hypothesis and also for the benefit of real world scenarios, we designed and ran two experiments for testing the robustness of the considered matching methods against disguises and random (and fixed) occlusions. This can also serve as quantitative measures to evaluate the ability of a method to bridge the modality gap. We stress that a method with high matching accuracy could rely on small/irrelevant parts of the image, weakening the robustness of the method and its generalization ability.

\subsection{Robustness to Disguises}\label{sec_sec_disguise}

\begin{figure}
  \centering
  \includegraphics[width=0.9\columnwidth]{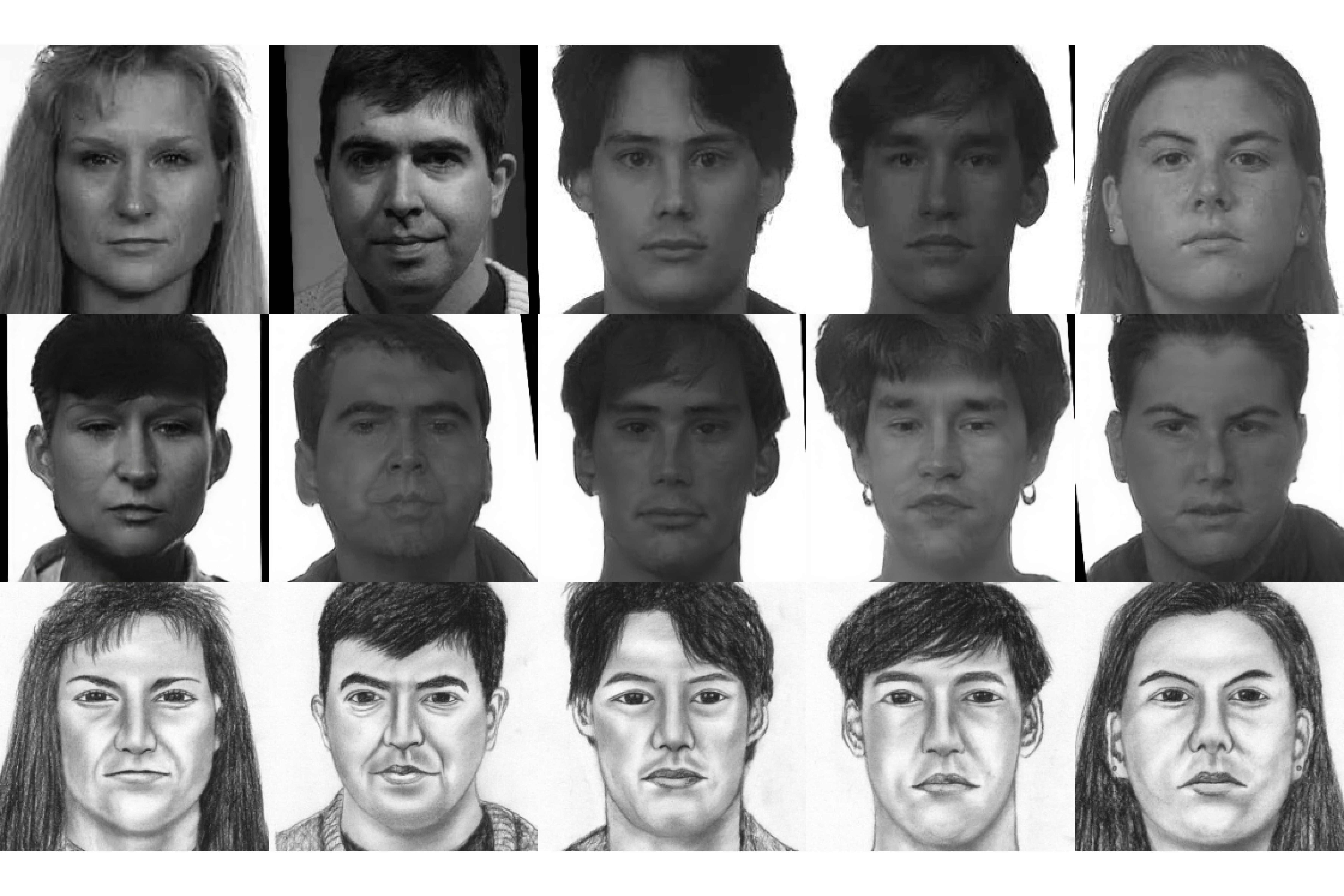}
  \caption{Examples of photos with disguises created based on CUFSF. We simulated the disguises by replacing the non-facial region of an face photo by that of another randomly-picked person. Top row: photos. Middle row: disguised photos. Bottom row: corresponding sketches.}
    \label{fig_face_disguise_examples}
\end{figure}

As discussed above, scenarios where suspects or persons of interest change their non-facial features especially hairstyles could occur in real-world law enforcement applications %
To simulate these scenarios, we replace the non-facial region of a person with the same region of anther randomly picked person. 
In other words, a disguised version of the face photo of person A is created by mixing A's facial region with a random person B's non-facial region. This has been realized using a face-swapping tool~\cite{SimSwap}. Examples can be found in Fig.~\ref{fig_face_disguise_examples}.

\subsubsection{Evaluation protocol}
We selected randomly 20\% of the test samples, denoted as ``d-set'', and swapped their non-facial features with one another such that each photo in the d-set has a new non-facial features. We then merge the d-set back with the rest (80\%) of the CUFSF dataset and readied a dataset for face-sketch matching with disguises. To evaluate the learned representations, we took the trained model as described in Section~\ref{subsec_hfr} without any retraining or fine-tuning and used it to match between disguised photos and sketches. 

\begin{figure}
  \centering
  \includegraphics[width=\columnwidth]{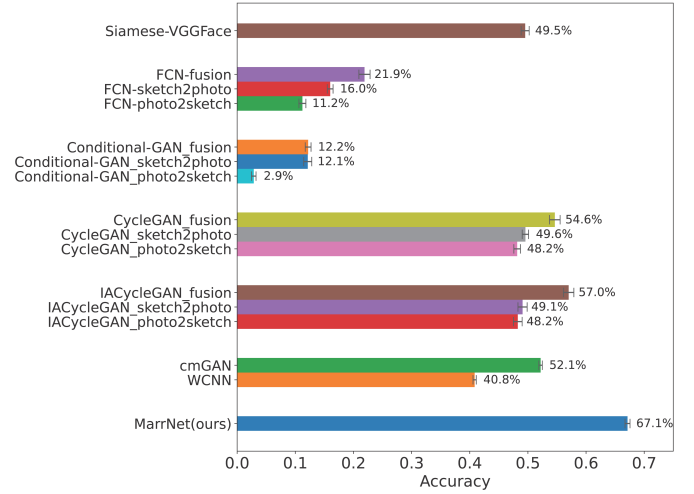}
  \caption{Matching accuracy between sketches and photos with simulated disguises. MarrNet outperformed existing methods by a large margin ($>$10\%).}
    \label{fig_match_acc_with_disguises}
\end{figure}

\begin{figure}
  \centering
  \includegraphics[width=0.9\columnwidth]{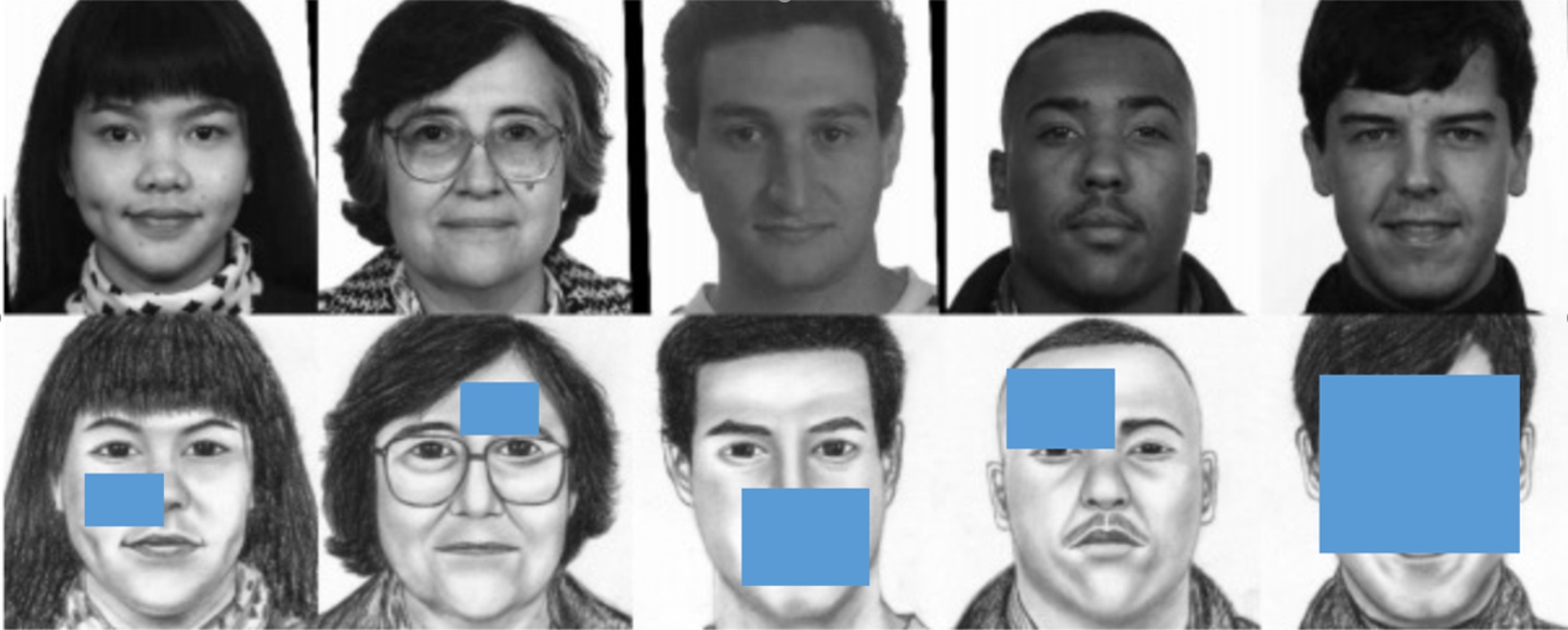}
  \caption{Examples of masked samples from the CUFSF dataset for the \textit{robustness to occlusions} test. Note that pixels in the masked regions were set to 255, but shown in blue here for the illustration purpose. }
    \label{fig_masked_test_examples}
\end{figure}

\subsubsection{Results and Analysis}
The averaged matching accuracy for disguised photos is reported in Fig.~\ref{fig_match_acc_with_disguises}. It can be seen that MarrNet outperformed the other methods by a large margin ($>$10\%), which demonstrated that the representations learned by MarrNet are much more robust than those learned by other methods.

Notably, Conditional-GAN's performance was extremely poor and even three times worse than the baseline. This indicated that poor quality modality transfer disrupted the subsequent feature learning process and led to wrongly-focused representations.

\subsection{Robustness to Occlusions}
We generalize the scenario of subjects of interest trying to hide their faces with disguises to testing the robustness of the matching methods to random occlusions. These tests were carried out on both CUFSF and cmRRUFF datasets. We took trained models as specified in the section~\ref{subsec_cmvsm} and 
\ref{subsec_hfr}, and tested their ability of matching samples with occlusions. 

\subsubsection{The compared methods}
Besides the considered methods, an additional method dubbed within-modality-matching was included to establish a baseline showing how masking itself affects the matching accuracy. To be specific, we performed within-modality matching with masking for each modality and averaged the matching accuracy. we performed face-face and sketch-sketch matching with masking using VGGFace. For Raman-infrared spectrum matching, we performed Raman-Raman and infrared-infrared matching using LeNet. More details can be found in Appendix~\ref{app_sec_wmmwm}.

\begin{figure}
  \centering
  \subfigure[On CUFSF]{
  \includegraphics[width=0.9\columnwidth]{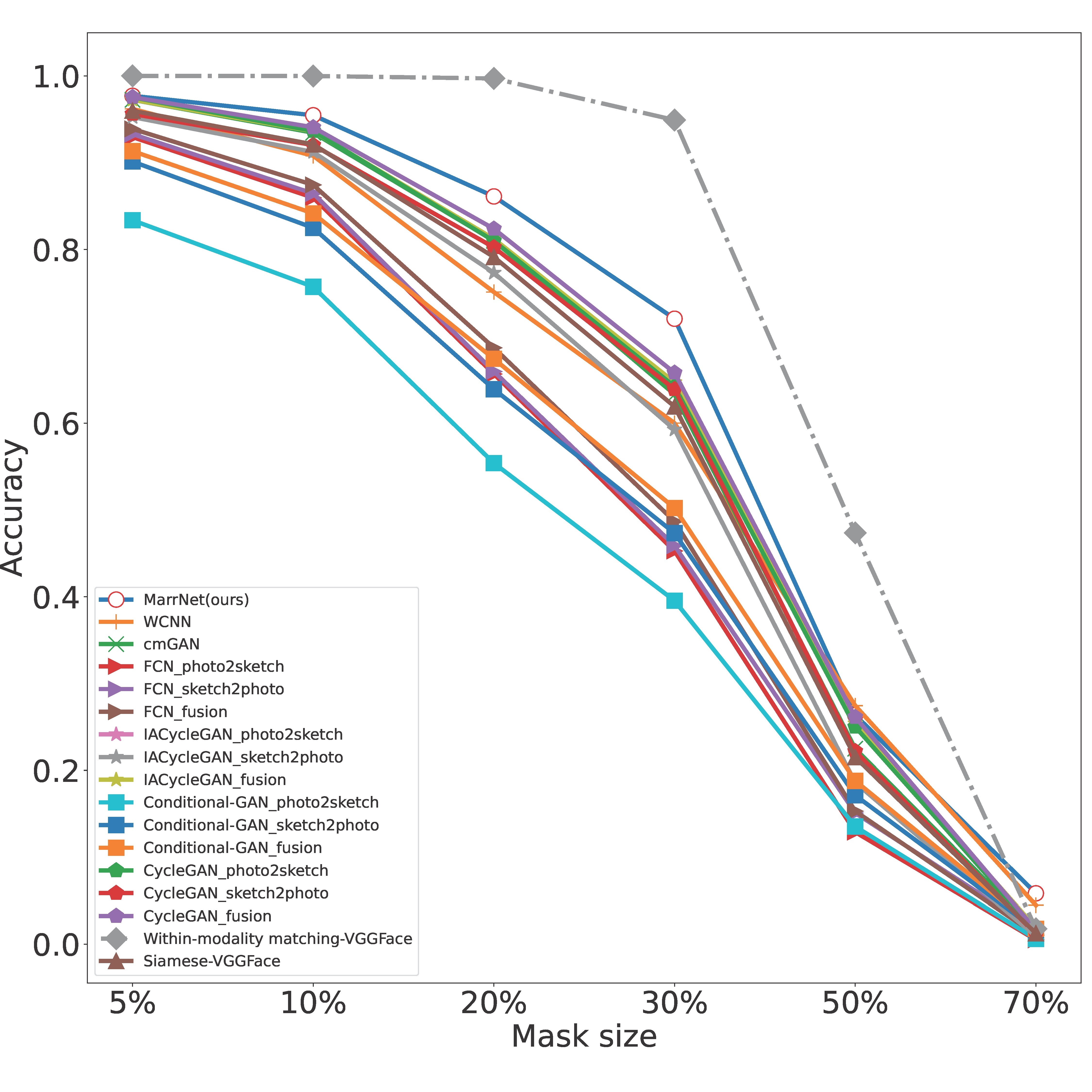}
  }
  \subfigure[On cmRRUFF]{
  \includegraphics[width=0.9\columnwidth]{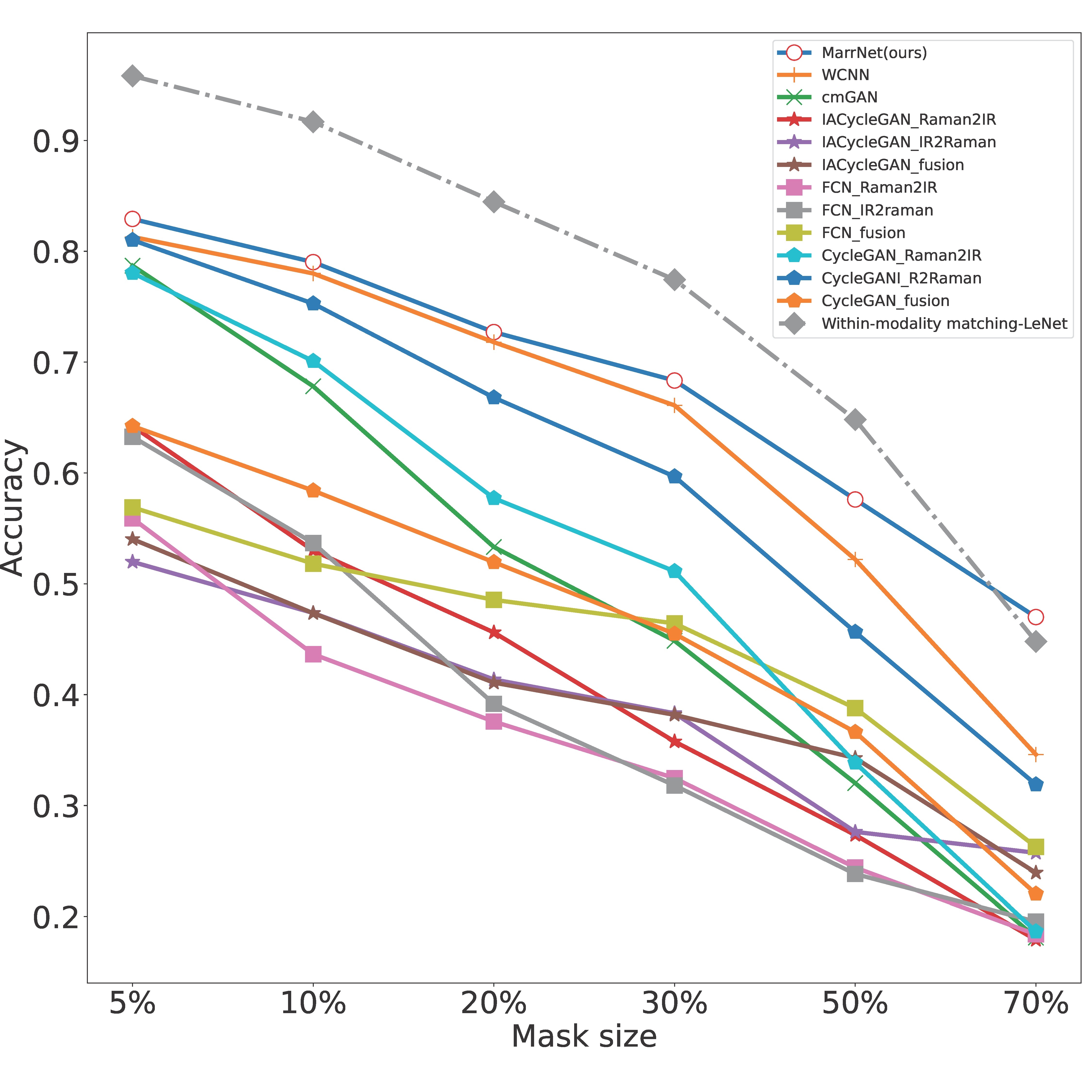}
  }
  \caption{Robustness to random occlusions of the compared methods on the CUFSF and cmRRUFF datasets. The x-axis is the size of random masks. The y-axis is the matching accuracy between photos and sketches with occlusions. Higher accuracy indicates better robustness to occlusions.}
    \label{fig_mask_face}
\end{figure}

\begin{figure}
  \centering
  \includegraphics[width=\columnwidth]{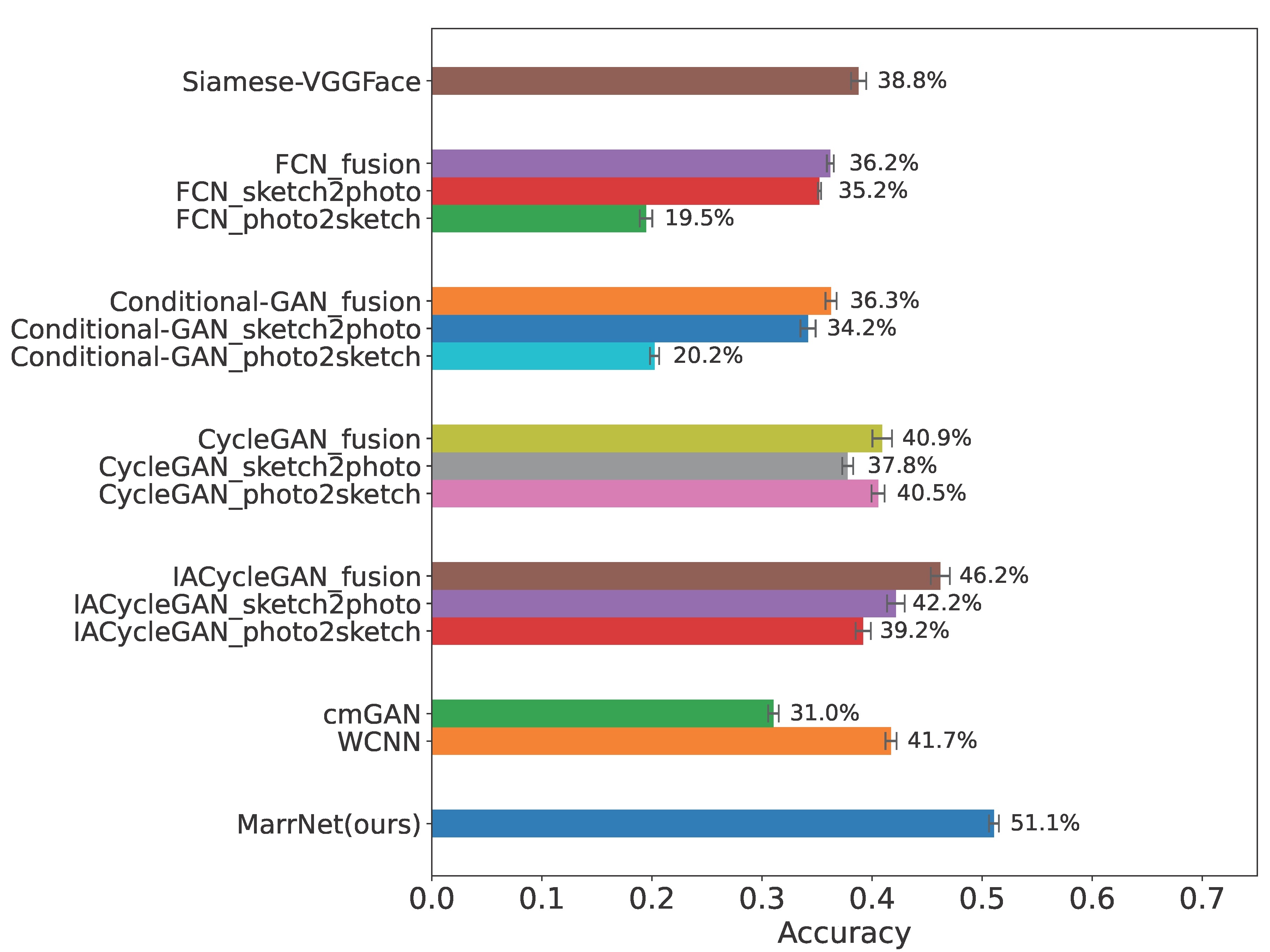}
  \caption{Robustness to fixed occlusions (facial masks) of the compared methods. MarrNet showed superior performance compared to existing methods.}
    \label{fig_match_acc_fixed_masks}
\end{figure}

\subsubsection{Evaluation protocol}
For heterogeneous face recognition on CUFSF, for each photo or sketch in the test set we generated $N=100$ masked samples where the masks are of square shape and placed them at random positions in the photo or sketch. The trained model was then used to match these masked samples against the library of sketches or photos, and the averaged matching accuracy was reported. Examples of masked samples can be seen in Fig.~\ref{fig_masked_test_examples}. The mask size was defined as the ratio of the length of masks and that of images. Note that for each tested method, we included only the pairs of test samples that were successfully matched in the robustness test, namely, the maximum matching accuracy of the robustness test on these pairs was 100\%. For Raman-infrared spectrum matching, the same protocol has been followed except that the signals and masks are one-dimensional.

\subsubsection{Results and Analysis}
Fig.~\ref{fig_mask_face} showed the results of the robustness to occlusions tests on the CUFSF and cmRRUFF datasets respectively. First of all, the matching accuracy of all methods decreased when the size of masks increased. This is expected, as larger masks on samples destroy more discriminant information. While for all mask sizes the proposed MarrNet exhibited substantially better performance than the existing state-of-the-art methods including IACycleGAN on both datasets. For heterogeneous face recognition, MarrNet outperformed existing methods such as WCNN and IACycleGAN by more than 10\% at the mask size of 30\%. For vibrational spectrum matching, the improvement was more than 20\% when the mask size was 70\%. It is worth noting that on CUFSF, IACycleGAN performed slightly better than the baseline, while significantly worse on cmRRUFF.

Note that most existing methods with explicit modality transfer performed better than the baseline method~\footnote{We referred to Siamese-VGGFace for face-sketch matching, Siamese-LeNet for Raman-infrared spectrum matching} on the dataset CUFSF, but worse on cmRRUFF. This suggested that these methods highly depended on the success of the modality transfer. For these methods, it is not possible to train an accurate matching system without quality transformed samples. %

Lastly, it is interesting to see that with large masks of 70\%, MarrNet performed even better than the within-modality matching, which indicates that the cross-modality representations learned by MarrNet are even better than the within-modality representations learned by VGGFace or LeNet when dealing with large occlusions.

To simulate subjects wearing facial masks, we have also conducted experiments of robustness to fixed occlusions (facial masks). Results can be found in Fig.~\ref{fig_match_acc_fixed_masks}. It can be seen that our MarrNet again outperformed the compared methods significantly.

\section{Conclusion and Future work}

We propose a cross-modality UNet module (cmUNet) to learn modality-agnostic representations by performing cross-modality transformation and in-modality reconstruction guided by reconstruction and adversarial losses. By combining this cmUNet with a standard Siamese feature extractor, we proposed MarrNet for cross-modality matching which learns modality-agnostic embedding and matching in an integrated manner. In five cross-modality recognition tasks, namely vibrational spectrum matching, heterogeneous (photo-sketch, visible-near infrared, visible-thermal) face recognition and visible-thermal person re-identification, MarrNet showed superior performance compared with existing state-of-the-art methods. Additional experiments on robustness to disguises and occlusions as well as analysis of saliency maps showed that MarrNet excelled in capturing the true discriminant information and learning robust representations which might be owing to it bridging the modality gap well. The proposed cmUNet can be straightforwardly applied to more than two modalities or fusing data from multiple different sources (tasks, modalities etc.) and we will investigate them in future work. Besides, we will also investigate the auto-learning of weights of the loss terms to better integrate the cmUNet to downstream neural modules.

\section*{Acknowledgment}
This research was supported in part by National Natural Science Foundation of China under grant No. 62076140, No. 62233011 and in part by United States -- Israel  Binational Foundation under grant (No. 2022641 to M.O.).

\bibliographystyle{ieeetr}
\bibliography{cmm_tip}

\appendix
\subsection{Implementation Details of the Compared Methods}
\subsubsection{Within-modality matching with masking}\label{app_sec_wmmwm}

To establish a baseline showing how masking itself affects the matching accuracy, we included a method \textit{within-modality matching} for comparison. The details of the method are as follows: assume that all the photos form a set $\mathbb{P}$, and for a photo $\textbf{x}_{p}$ and the corresponding masked version $\Tilde{\textbf{x}}_{p}$, the accuracy of the within-modality matching is calculated as 

\begin{align}
    \label{eq_within_modality_matching}
    \Big (\sum_{x_{p} \in \mathbb{P}}  \big( \textbf{x}_{p} == \mathop{\arg\min}\limits_{x \in \mathbb{P}} \| F(\textbf{x}) - F(\Tilde{\textbf{x}}_{p}) \|_{2} \big) \Big) / |\mathbb{P}|
\end{align}

where $F$ is a feature extractor to calculate embeddings. VGGFace and LeNet were chosen to be $F$ for two applications respectively which were included for comparison as shown in Table~\ref{tab_result_CUFSF} and Table~\ref{tab_result_RRUFF}.

Following (\ref{eq_within_modality_matching}), we performed face-face(masked) and sketch-sketch(masked) matching respectively on CUFSF and reported the averaged matching accuracy in Fig.~\ref{fig_mask_face}(a). Similarly, we tested the within-modality matching on RRUFF and reported the results in Fig.~\ref{fig_mask_face}(b).

\subsubsection{FCN}
Fully convolutional neural networks (FCNs) have been used in photo-sketch generation in previous work e.g.~\cite{zhang2015end,FCN2017}, we included FCN for comparison for face-sketch matching. We have also found that it worked reasonably well on Raman-IR spectrum matching and outperformed all the other methods except for the proposed MarrNet, see Table~\ref{tab_result_RRUFF}.

\end{document}